\documentclass[lettersize,journal]{IEEEtran}
\usepackage{amsmath,amsfonts}
\usepackage[ruled,vlined]{algorithm2e}
\usepackage{xcolor}
\usepackage{array}
\usepackage{multirow}
\usepackage[caption=false,font=normalsize,labelfont=sf,textfont=sf]{subfig}
\usepackage{textcomp}
\usepackage{stfloats}
\usepackage{url}
\usepackage{verbatim}
\usepackage{graphicx}
\usepackage{cite}
\usepackage{caption} 
\usepackage{comment}
\usepackage{orcidlink}
\begin{document}

\title{Automatic counting of planting microsites via local visual detection and global count estimation}
\author{Ahmed~Zgaren, \hspace{1cm}
        Wassim~Bouachir, \hspace{1cm} and 
        Nizar~Bouguila % <-this % stops a space
}
\maketitle


\begin{abstract}
In forest industry, mechanical site preparation by mounding is widely used prior to planting operations. One of the main problems when planning planting operations is the difficulty in estimating the number of mounds present on a planting block, as their number may greatly vary depending on site characteristics. This estimation is often carried out through field surveys by several forestry workers. However, this procedure is prone to error and slowness. Motivated by recent advances in UAV imagery and artificial intelligence, we propose a fully automated framework to estimate the number of mounds on a planting block. Using computer vision and machine learning, we formulate the counting task as a supervised learning problem using two prediction models. A local detection model is firstly used to detect visible mounds based on deep features, while a global prediction function is subsequently applied to provide a final estimation based on block-level features. To evaluate the proposed method, we constructed a challenging UAV dataset representing several plantation blocks with different characteristics. The performed experiments demonstrated the robustness of the proposed method, which outperforms manual methods in precision, while significantly reducing time and cost.
\end{abstract}

\begin{IEEEkeywords}
Object counting, Object detection, computer vision, precision forestry, UAV imagery.
\end{IEEEkeywords}

\section{Introduction}
\begin{figure}
   \centering
\subfloat[][]{\includegraphics[width=0.24\textwidth]{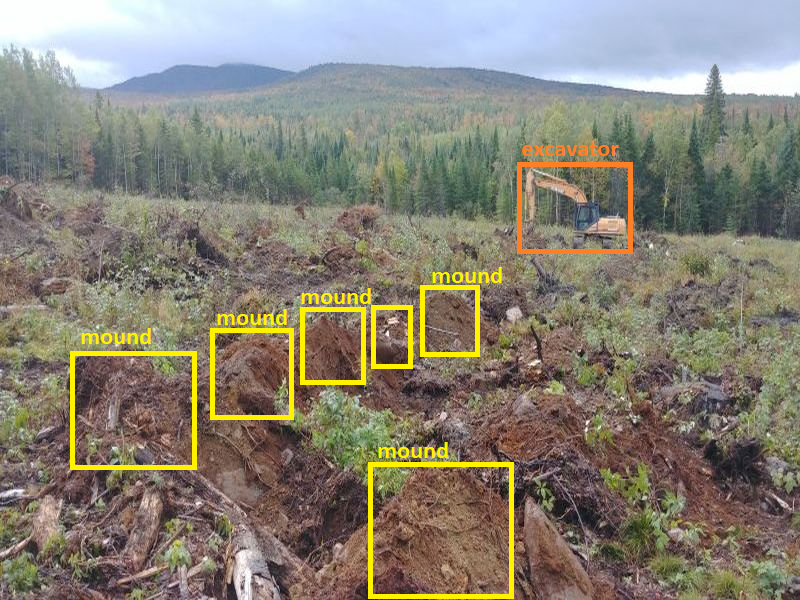}}%
\hfil
\subfloat[][]{\includegraphics[width=0.24\textwidth]{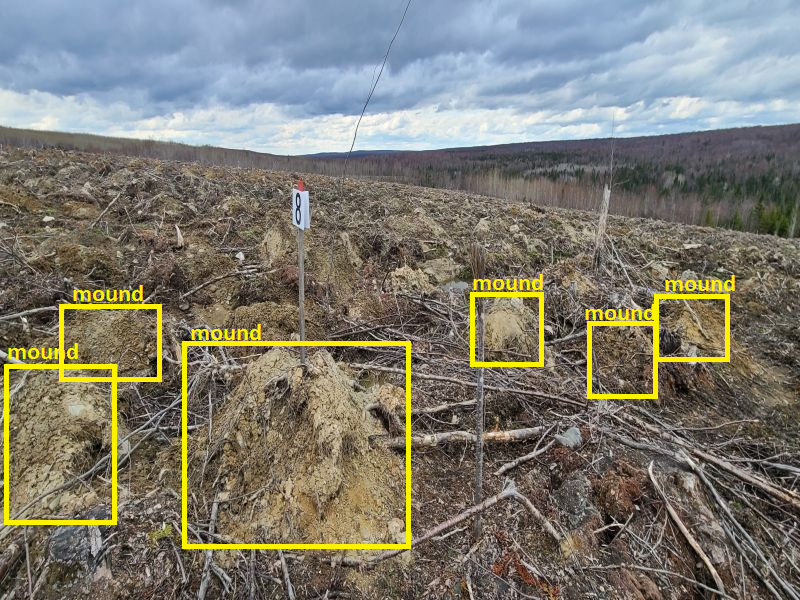}}%
\hfil

    \caption{Examples of mechanically prepared mounds in the balsam
fir-white birch bioclimatic domain in Quebec, Canada. The fields present irregularities between different planting blocks, as well as between regions within the same planting block. The created mounds have different appearances, shapes, and sizes.}
    \label{fig:mound_example}
\end{figure}

\IEEEPARstart{M}{echanical} site preparation by mounding is a popular technique often used by forest managers to ensure optimal growth conditions for tree seedlings. This procedure allows to address several soil problems that can severely impede root establishment of the planted trees, including high soil bulk density due to compaction by logging machinery, high water table, and plant competition for resources  \cite{Wolken2010DifferencesII}. Mechanical preparation for planting is performed by using machinery (e.g. excavator) to create a mound corresponding to each planting microsite (see figure \ref{fig:mound_example}). Prior to planting operations, the number of planting microsites (mounds) present should be estimated, in order to determine the number of tree seedlings to be planted on each site. 

The number of tree seedlings is often estimated by manual count, where several workers move through the field and visually identify each prepared mound on a portion of the site. The number of mounds is then estimated for the entire site by interpolating the partial result and assuming a constant density of mounds. This traditional counting method is time consuming and subject to errors. Moreover, the interpolation technique is not always reliable, as the density of mounds varies due to several factors, including site characteristics, preparation quality and type of machinery used.
\IEEEpubidadjcol  
With the increased use of drones in the forest industry, some forest managers attempted to replace field work by photo-interpretation of UAV images. In this context, photo-interpretation consists of the detection of mounds by a human operator on a portion of the UAV image. Similarly to the fieldwork procedure described above, the final estimation is then predicted for neighboring regions, by interpolating the result. Although this method allows to significantly reduce the time required for field work, photo-interpretation is also subject to human error, in addition to imprecision due to terrain variability factors. These estimation errors often result in complex and imprecise handling of seedlings in the field, which causes
monetary losses and additional planting delays. 
%This method decreases the count time but still have an error superior to 10\% due to terrains variability and non homogenieties between blocks in the same terrain.  
%Mounding occurs during site preparation stage (see \ref{fig:mound_example}) and needs to accurately estimates the number of mounds for next stages. However, and due to complexity of ground properties, the number of mechanically prepared sites varies from one site to another. Traditional counting is done manually by several workers which .\\
%A high error in estimating mounds number causes significant delays in planting and to huge loss of money.  With the growth of the incorporation of Unmanned Aerial Vehicle (UAV), which is a new remote sensing platform, flexible, cost-efficient, and remotely operated by radio technologies, UAV's have replaced satellite images on geoscience field, and earth imagery for 
%forest industry and applications \cite{Oh2020PlantCO, Gu2020ACO, Windrim2019AutomatedMO}, UAVs are easily integrated in a wide range of applications such as surveillance \cite{Chriki2020}, logistic \cite{Villa2020ASO}, environmental \cite{Fassnacht2016ReviewOS}, and  forestry where coupled with computer vision algorithms for automated tasks like tree detection \cite{Safonova2019DetectionOF, Osco2020ACN}, fire detection \cite{Shin2019UsingUM}, and tree counting \cite{Hassaan2016PrecisionFT}.\\

%The rise of remote sensing engines led to an increase of deep learning algorithms usage. In fact, these methods enhanced the performance of results in many applications which attracted both the academic and industrial communities to integrate them in a wide range of applications \cite{Chriki2020, Villa2020ASO, Fassnacht2016ReviewOS}. An important application in the computer vision field is object counting in crowded scenes.

% Recently, the immersion of artificial intelligence with UAVs opened new areas of research for new computer vision applications, with the objective of automating the analysis of UAV images.  Examples species identfication \cite{Baena2017IdentifyingSF} and trees \cite{Nezami2020TreeSC} evolution and count, among several others.

 Our work aims to propose an automatic process to fully take advantage of UAV imagery, coupled with computer vision techniques. We formulate the task of mound counting as a supervised learning problem based on the combination of two prediction models: 1) a local detection model used for detecting mounds based on deep features, and 2) a global estimation model for predicting the final count based on global features. 
  The proposed system is trained in an offline manner to resolve two different, yet related, problems consecutively. On the one hand, the base object detector (local model)  is trained to recognize visible mounds using a large dataset of manually annotated UAV images. On the other hand, the global estimation function is learned to predict the final count based on global information on the target field, including the detection result. Once the system is trained, online counting of mounds on a new image is performed by applying local and global models sequentially. More specifically, counting by detection is firstly performed using the local visual detector to provide a preliminary counting of visible mounds. The preliminary count is then fused with global information extracted from the block's orthomosaic, to form a global feature vector. This information is finally used as the input of the global estimation function to obtain a precise prediction of mound count.

The proposed framework allows to automate the counting process by taking advantage of computer vision and UAV imagery. On the one hand, the use of our framework significantly reduces counting time, which currently takes several days to be accomplished using the manual method. On the other hand, we also improve counting precision, which has the potential to provide substantial financial benefits to forest managers.
  %
% we demonstrate that two models are essential to... erosion/deterioration caused by heavy rain events. (c) occlusion due to the presence
%of woody debris on the forest floor

We performed extensive experiments by testing the proposed method on UAV orthomosaics captured by overflying 12 planting blocks with different characteristics. These blocks are not included in the training set and contain 6164 mounds on average per block. The performed experiments demonstrate the high accuracy and robustness of the proposed method. With an overall relative counting precision of 95\%, our computer vision solution outperforms both the manual field method as well as the photo-interpretation method. Further, we present an ablation study to investigate the contributions of each system module and several design strategies. Our analysis demonstrates the importance of different system components. In particular, it underlines that the combination of the two models (global and local) is more efficient than only applying a local visual detector. This is mainly due to the nontrivial nature of our object detection problem, with a high appearance variability at the scene level, and where objects of interest could be invisible due to several perturbation factors (e.g. occlusion by woody debris, water accumulation, mound erosion, and destruction).

Our main contributions can be summarized as follows:
\begin{itemize}
    \item We propose a fully automated vision-based system to count mounds in orthomosaics. We thus contribute in addressing an important forestry management problem, by improving fieldwork conditions and significantly reducing time, money, and resource consumption for forest managers.
    \item We propose a robust computer vision framework, which can be generalized for a wide range of object counting applications, especially those using UAV imagery in crowded scenes.
    \item We construct a new dataset for mound detection and counting from UAV-based imagery.
    \item We analyse of the visual mound detection problem to identify the main challenges to be addressed in future work.
\end{itemize}

This paper is structured as follows. Section \ref{RW}, reviews relevant works from the literature on automatic object counting. In section \ref{propMethod}, we present the proposed method for mound detection and counting. Comprehensive experimental work is presented in section \ref{experim}, including system evaluation in real-world conditions as well as an ablation study. In section \ref{disc}, we discuss the performance of our system with respect to several challenging situations. Finally, section \ref{concl} concludes the paper.

\section{Related works}\label{RW}

The use of aerial imagery stimulated research work in several application areas related to forestry and agriculture, such as tree detection \cite{Santos2019AssessmentOC}, animal counting \cite{Chamoso2014UAVsAT}, tree species classification \cite{Baena2017IdentifyingSF}, biomass estimation \cite{Gennaro2020AnAU}, and fire monitoring \cite{Shin2019UsingUM}. Several of these applications are based on the task of estimating the number of objects present in images, referred to as crowd counting. This section reviews the most important studies and methods proposed to count objects in crowded scenes. Methods from the literature can be categorized into three main approaches: 1) the traditional approach, mostly based on hand-crafted features and classical machine learning models, 2) the Deep Learning (DL) approach, where DL models are applied to learn and classify crowd regions, and 3) hybrid approaches, where both hand-crafted and deep features are used to improve the overall performance.

\subsection{Crowd counting using traditional approaches}

Traditional approaches are mostly based on image processing techniques to extract hand-crafted features \cite{Velastin1993AnalysisOC,Regazzoni1993ARV, Davies1995CrowdMU, Regazzoni1996DistributedDF}, in addition to machine learning methods to predict object count. We can distinguish three main categories of methods: detection-based methods, where the final count is the total number of detected objects, regression-based methods, where a regression function maps the final count to the input features, and density estimation-based methods,  where the final count is extracted from an estimated density map.

\subsubsection{Detection-based methods}

These methods generally adopt detection frameworks \cite{Dalal2005HistogramsOO, Viola2004RobustRF, Wu2006DetectionAT, Sabzmeydani2007DetectingPB} by training a machine learning classifier on hand-crafted features. The used features include Harr wavelet \cite{Viola2004RobustRF}, histogram of oriented gradient (HOG) \cite{Dalal2005HistogramsOO}, edgelet \cite{Wu2006DetectionAT} and shapelet \cite{Sabzmeydani2007DetectingPB}. Detection is then performed either by global \cite{Dalal2005HistogramsOO, Leibe2005PedestrianDI, Tuzel2008PedestrianDV, Enzweiler2009MonocularPD}, part-based \cite{Wu2006DetectionAT, Felzenszwalb2009ObjectDW, Lin2001EstimationON}, or shape learning \cite{Zhao2008SegmentationAT} techniques. Several nonlinear classifiers have been also used to learn object patterns, such as support vector machine (SVM) \cite{Ilao2018CrowdEU}, boosted trees \cite{Zhou2016CountingPU} and random forests \cite{Pham2015COUNTFC}.

methods for counting by detection are generally successful in moderately crowded scenes but fail in handling highly-crowded scenes due to occlusions and scene clutter. 

\subsubsection{Regression methods}

To overcome the dependency on learning detectors, regression-based methods aim to map the final count to input local features \cite{Chan2009BayesianPR, Ryan2009CrowdCU, Chen2012FeatureMF}. Models in this category are designed based on two main steps: low-level feature extraction and regression modeling.

Various types of features such as foreground features, edge features, texture, and gradient features have been used to extract visual information. While background subtraction techniques have been used to separate foreground features, Blob-based holistic features such as area and perimeter have also been successfully used to capture global properties of the scene \cite{Ryan2009CrowdCU, Chen2012FeatureMF, Chan2008PrivacyPC}. To further increase accuracy, more advanced local features have been used, such as edges \cite{Mikolajczyk2003ShapeRW} and texture/gradient features \cite{Tceryan1993TextureA, Hwang2004AdaptiveII}.

Once feature extraction is performed, different models can be used to learn a mapping function from low-level features to crowd count, such as linear regression \cite{Paragios2001AMA}, piece-wise linear regression \cite{Chan2008PrivacyPC}, and neural networks \cite{Marana1998OnTE}.

Idrinet al. \cite{Idrees2013MultisourceMC} observed that there is no single feature or detector which is capable of providing sufficient information for precise counting in high-density scenes. They proposed to combine different feature extraction methods to capture multiple information types. Once local counts in all image patches are collected, they are fed to a global multiscale Markov Random Field (MRF) to estimate the final count.

Counting by regression mainly attempts to mitigate the former problems, like occlusion and scale variation, while improving crowd counting performance in highly crowded scenes. However, low-level feature representations are often unable to capture semantic information, and regressors tend to ignore the spatial information by regressing global count.

\subsubsection{Density estimation-based methods}

To handle the spatial information problem, Lempitsky et al. \cite{Lempitsky2010LearningTC} introduced a model to linearly map between local patch features and their corresponding object density maps. They formulated density map learning as a minimization of a regularized risk quadratic cost function using a new loss function for learning. Motivated by the success of density map estimation, Pham et al. \cite{Pham2015COUNTFC} proposed a nonlinear mapping model using random forest regression. They proposed a crowdedness prior to overcoming the problem of the large  variation in appearance and shape, then trained two different forests corresponding to this prior. Wang et al. \cite{Wang2016FastVO} proposed a faster method by clustering the feature space into subspaces and then by learning the embedding of each subspace formed by image patches. Xu et al. \cite{Xu2016CrowdDE} observed that crowd density estimation methods are based on computationally expensive Gaussian process regression or Ridge regression models, which can only handle a small number of features. They proposed to incorporate a rich set of features to boost the performance of crowd density estimation.

Crowd counting by traditional methods was successful in moderately crowded scenes and the proposed models did not require significant computational resources. However, traditional methods are based on low-level features for image representation, which limits their capability to handle difficult situations like appearance variation, scale variation, and context information. 
%Also, learning a non-linear function for mapping between count and local features is not efficient at accurate count estimation.

\subsection{Deep learning for Crowd counting}

During the last decade, deep learning models have outperformed traditional machine learning approaches in various application fields. Due to their deep and sophisticated architectures, DL models proved their ability to extract robust semantic information from large datasets, which allows handling multiple challenging situations such as scale change, rotation, and appearance variation.

Motivated by the success of deep neural networks in computer vision, researchers have attempted to solve the object counting problem in crowded scenes using DL models. These models are mostly based on Convolutional Neural Networks (CNNs) to learn a non-linear function from crowd images to their corresponding density maps or counts.

\subsubsection{Basic CNN approaches}

Since these methods are among the first deep learning formulations for crowd counting, they integrate basic CNN layers to the proposed models. Wang et al. \cite{Wang2015DeepPC} proposed an end-to-end CNN regressor for people counting. The model is based on AlexNet \cite{Krizhevsky2012ImageNetCW} architecture, where the final fully connected layer is replaced by a single neuron to predict the final count. Fu et al. \cite{Fu2015FastCD} optimized a multistage CNN  by removing some network connections according to the observation of the existence of similar feature maps. Then, two CNN cascade classifiers were designed to classify crowd images into five classes: very low, low, medium, high, and very high. To optimize the model on cross-scene crowd counting, Zhang et al. \cite{Zhang2015CrosssceneCC} proposed to alternatively train a CNN for two tasks: density estimation and crowd counting. To further improve results on a target scene, the model is fine-tuned using training samples similar to the target. To increase counting accuracy and time processing, Walach et al. \cite{Walach2016LearningTC} proposed an improved CNN architecture with layered boosting and selective sampling techniques during training. In \cite{Shang2016EndtoendCC}, the entire image was fed to the network for count prediction, instead of dividing it into patches. The authors combine a CNN with a recurrent neural network (RNN) to take advantage of contextual information when predicting both local and global counts. 

\subsubsection{Scale-aware CNN approaches}

More advanced CNN architectures were demonstrated to be robust to scale variation, by incorporating different techniques such as multi-column or multi-resolution networks \cite{Ciresan2012MulticolumnDN}. To ensure robustness and scale invariance, Boominathan et al. \cite{Boominathan2016CrowdNetAD} proposed to combine deep and shallow fully convolutional networks to capture semantic informer density map estimation and to ensure robustness to non-uniform scale and perspective variations.
Zhang et al. \cite{Zhang2016SingleImageCC} presented the Multicolumn CNN (MCNN), which consists of three columns for different scale capture (large, medium, small). Onoro and Sastre \cite{OoroRubio2016TowardsPO} developed a scale-invariant CNN model (HydraCNN). Firstly, they introduced the Count CNN (CCNN) architecture that incorporates the perspective information for geometric correction of the input features. Then, they constructed a network of three heads and a body. Each head is a CCNN architecture for learning features of a particular scale. Finally, the outputs of all the heads are concatenated and fed to the body to estimate the final density map.

Sam et al. \cite{Sam2017SwitchingCN} proposed a scale-aware crowd counting model based on switching CNNs. Their architecture consists of multiple MCNN regressors and a switch classifier trained to select the optimal regressor.
Later, Ranjan et al. \cite{Ranjan2018IterativeCC} proposed a multi-stage crowd count process (ic-CNN) to generate high-resolution density maps. The model architecture is based on stacking multiple ic-CNN, which consists of a two-branch CNN, where the first branch produces a low-resolution density map and the second merge the outputs of the first branch to produce high-resolution density maps. In a different approach to incorporate scale information, Zhang et al. \cite{Zhang2018CrowdCV} extracted feature maps from different layers of a backbone and combined them to produce the final density map. In \cite{Jiang2019CrowdCA} Jiang et al. proposed a Trellis Encoder-Decoder (TEDnet) for crowd counting. The model focuses on generating high-quality density estimation maps. They modified the Trellis architecture to incorporate rich scale information by using dense skip connections across paths. Recently, Dong et al. \cite{Dong2020CrowdCB} proposed an end-to-end scale-aware model (MMNET) that integrates multi-scale features generated by different stages, to handle scale variations. 
Hu et al. \cite{Hu2020NASCountCW} estimate the density map using Neural Architecture Search (NAS), which exploits multi-scale features and addresses the scale variation issue in counting by density. Moreover, Liu et al. \cite{9765482} proposed a multiscale parallel encoder by introducing an Efficient and Lightweight Convolution Module (ELCM) to extract different scale features. Then, a Scale Regression Module (SRM) decoder is used to generate the density map. More recently, Aldhaheri et al. \cite{Aldhaheri2022MACCNM} proposed to filter out background noise from foreground features using a segmentation guided attention mechanism.

\subsubsection{Context-aware CNN approaches}

Incorporating local and global contextual information into the CNN architecture for crowd counting is a complex task that has attracted several researchers. Sheng et al. \cite{Sheng2018CrowdCV} proposed to integrate semantic information by learning locality-aware feature (LAF) sets. The proposed architecture comprises three main components. First, a CNN transforms the input raw pixel high-resolutions attribute feature map. Then, following the idea of spatial pyramids on neighboring patches, the LAF is introduced to explore more spatial context and local information. Finally, the local descriptors from adjacent cells are encoded into image representations using the VLAD \cite{Jgou2010AggregatingLD} encoding method. Moreover, Ilyas et al. \cite{Ilyas2019CASACrowdAC} have designed an end-to-end CNN model for crowd counting. The architecture of the proposed method consists of two parts: a deep feature extraction network (DFEN), and a scale-aggregation module with dilated convolution (SAD). They combined DFEN and SAD to collect large-scale contextual information, handling the perspective distortion and expanding the spatial sampling location. 
Also, Tian et al. \cite{8897143} combined Density Aware Network (DAN) and Feature Enhancement layer (FEL) to capture local and global contextual features. In a different training approach, Lei et al. \cite{Lei2021TowardsUC} proposed a weakly-supervised density estimation model. The model was trained using primary and auxiliary brands and two different annotation types, location level annotation, and count level annotated images. Recently, Do \cite{9701500} proposed to learn global context information using a visual transformer. Liu et al. \cite{9807329} proposed a multi-task Encoder-Decoder density map generator to learn the counting of maize stand from UAV images. To alleviate the problem of density dependence, the authors proposed to incorporate count and density map errors into the loss function.

Crowd counting using CNN-based architecture has been successful in both moderately and highly crowded scenes. Following this direction, multiple network architectures have been introduced to handle challenging situations such as scale variations and occlusion. %However, the count accuracy was not precise due to density estimation approach which ignores local and contextual information due to convolution operator used at the feature extraction level and the integral function used for count estimation.
\subsection{Hybrid methods}
Hybrid methods mainly combine two feature representations: hand-crafted features, and deep features. Features can be extracted explicitly or implicitly, to be used to train either a classical machine learning classifier or a deep neural network to count objects. Lin et al. \cite{Lin2021AnEF} proposed a method for counting persons in videos when crossing a line for low-cost devices. The proposed architecture uses a knowledge distillation approach to transfer knowledge from the CNN object detector, in order to train a small Local Binary Patterns (LBP) cascade classifier. After YOLO \cite{Redmon2018YOLOv3AI} detects persons in video frames, images of persons with confidence higher than 30\% are used to train the LBP cascade classifier, which is used to detect-and-track pedestrians.   
Bouachir et al. \cite{Bouachir2019ComputerVS} proposed a two-stage detector to automate the estimation of the number of planting microsites from multispectral images. The first stage is a cascade detector based on LBP features to generate region proposals that are likely to correspond to objects of interest. The second stage consists of a CNN applied for candidate region classification. This method allowed to handle difficult situations where the objects are difficult to detect (e. g. appearance variability, motion blur). However, this method suffers from a high computational cost due to the use of a sliding window process and region proposals.  Besides, hybrid methods in general may also suffer from high computational complexity due to the fusion of multiple features and the use of more than a single classifier.

In our work, we propose an hybrid framework where we adopt two different approaches sequentially. However, unlike the above discussed hybrid methods that often suffer from high computational complexity, we perform visual feature extraction only during the first detection stage. This allows to maintain a reasonable computational complexity that corresponds to that of the object detector. Further, our second prediction stage is inspired by regression methods in that counting is predicted globally, regardless of individual object localization. Nevertheless, we do not process any visual low-level features, as our regression function is trained on global properties describing an orthomosaic. As demonstrated further in our experiments, the proposed conception allows to significantly improve counting precision compared to merely using a state-of-the-art object detector.

%To construct the detector and validate our method, a new dataset of high resolution RGB images was captured using UAV on different sites. A discussion of the challenges and recommendations for further works will be presented by the end of this study.

\begin{figure*}
    \centering
    \includegraphics[width = 0.9\textwidth]{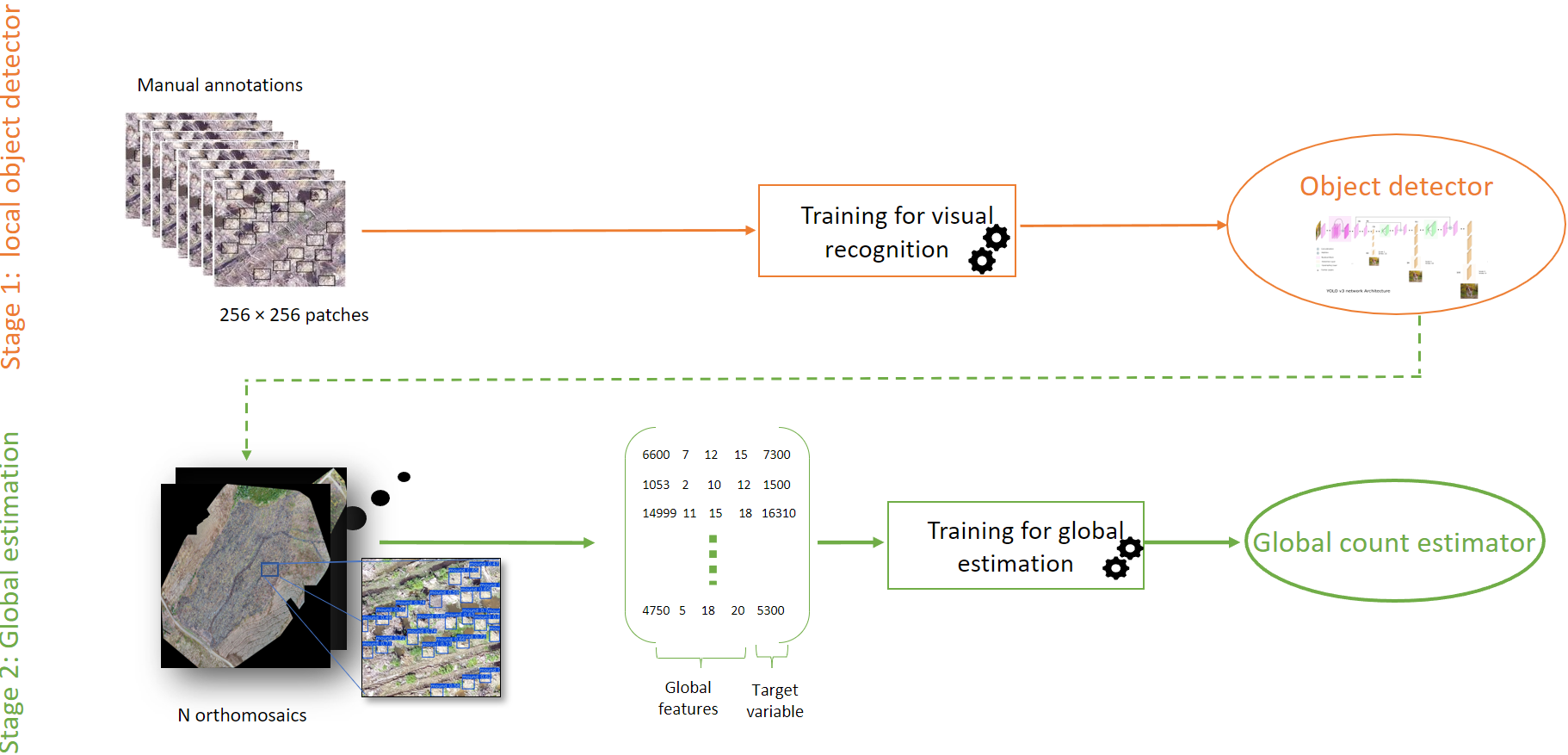}
    \caption{System training. \textit{Stage 1} illustrates the visual detector construction and \textit{Stage 2} the global correction module training. Continued arrows indicate the transition between steps and dashed arrows indicate the use of a system component in the target step.}
    \label{fig:trainfig}
\end{figure*}

\section{Proposed method}\label{propMethod}
\subsection{Motivations and overview}
% justify: 1 features, 2 detection model, 3 correction step, etc wrt related works 

\begin{figure*}
   \centering
\subfloat[][]{\includegraphics[width=0.33\textwidth]{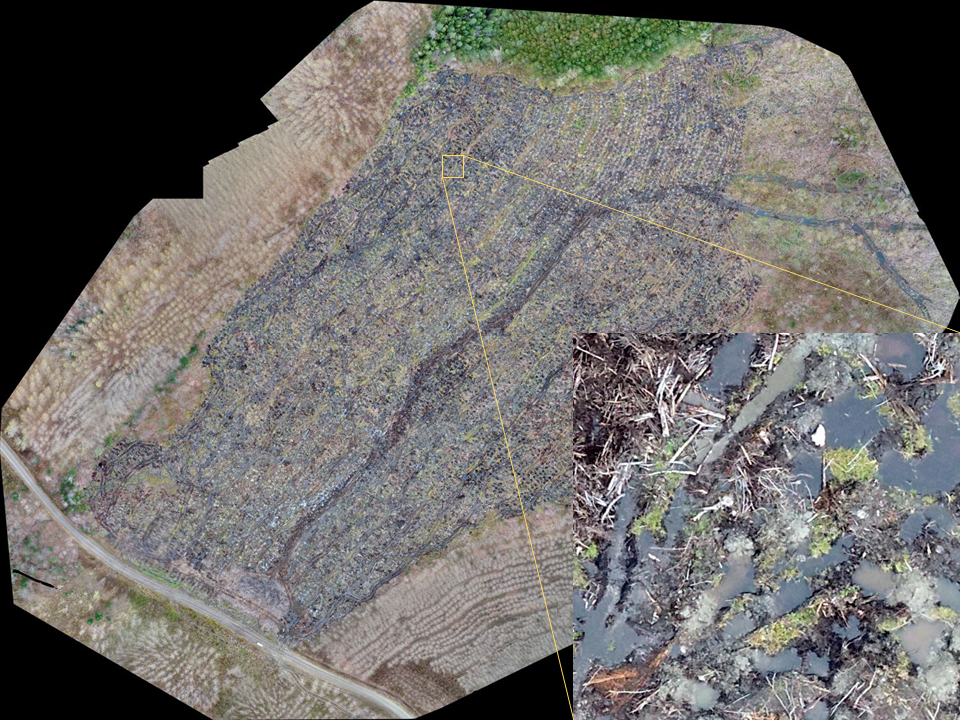}}%
\hfil
\subfloat[][]{\includegraphics[width=0.32\textwidth]{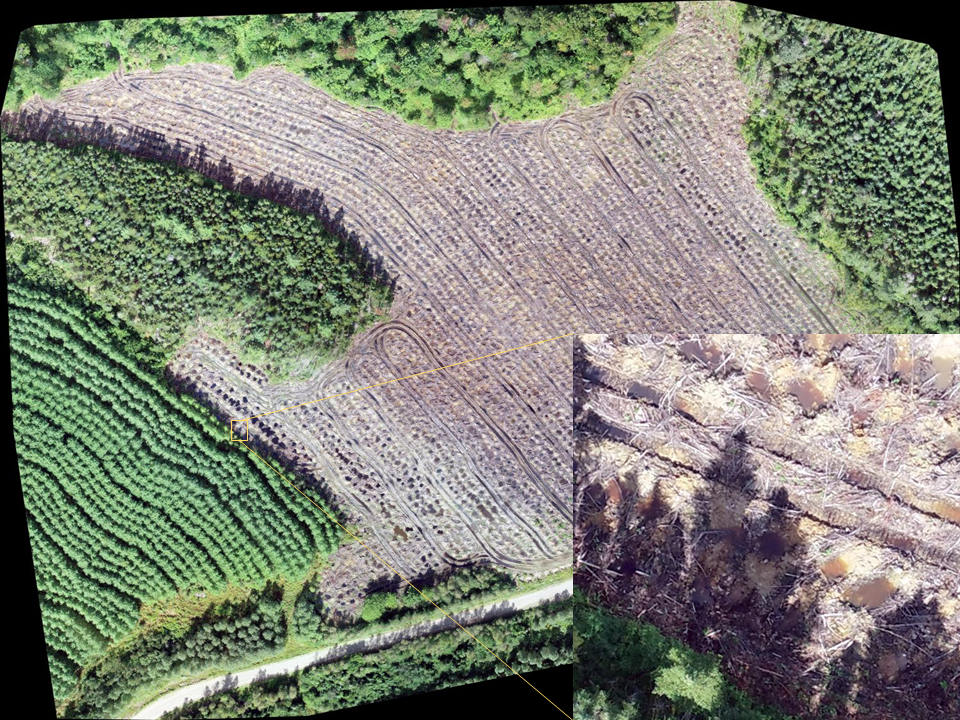}}%
\hfil
\subfloat[][]{\includegraphics[width=0.32\textwidth]{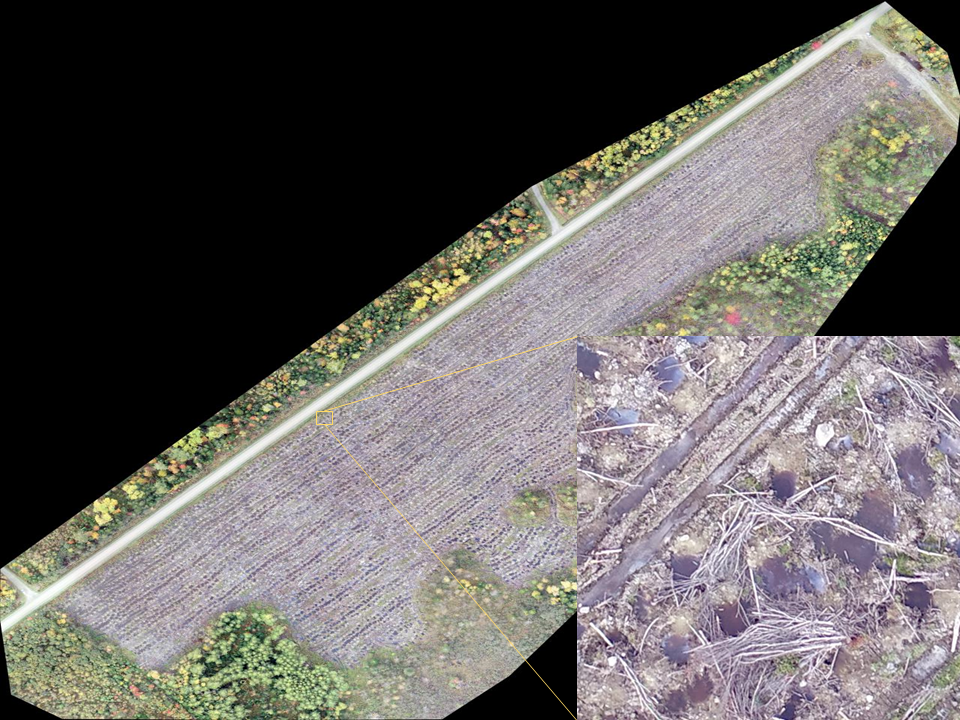}}%
\hfil
    \caption{Examples of orthomosaics captured and reconstructed for 3 different planting blocks. Corresponding areas are 13 hectares (ha), 4 ha, and 6 ha, respectively for (a), (b), and (c).}
    \label{fig:ortho}
\end{figure*}

The aim of our work is to design a method to accurately estimate the number of planting microsites (mounds) present on mechanically prepared planting blocks. The input of the proposed system is an orthomosaic representing a new planting block (as shown in figure \ref{fig:ortho}), each orthomosaic being constructed from high-resolution UAV images. We can also see in figure \ref{fig:ortho} that visual recognition of mounds could be a complex task even for a human. This is mainly due to several challenges, such as the similarity in appearance between mounds (objects of interest) and surrounding terrain (background), shape/appearance variation between mounds (intra-block variability), and appearance variability between different planting blocks (inter-block variability).  

To handle these difficulties, system training is done by constructing two prediction models consecutively. We first construct a visual object detector from manual annotations, to locally recognize visible mounds individually. Deep features are used in this first model for the visual representation of objects. In fact, hand-crafted features have been demonstrated to be limited for such a complex detection problem, which may result in a large number of false positives \cite{Bouachir2019ComputerVS}. Further, we construct a global estimation model for predicting the number of mounds from the global features of a given planting block. The used feature set comprises block-level information, such as block area and prior knowledge on mound density. The system training process is illustrated in figure \ref{fig:trainfig}.

Once the entire system is trained, mound counting on an orthomosaic representing a new planting block is performed by applying the two models subsequently. That is, the local detection model is first applied to perform counting-by-detection, which is considered as a preliminary estimation of the number of mounds based on visual recognition. Note that local detection is preceded by a fine-tuning process to handle the inter-block variability problem mentioned above. This is achieved by providing the detection model with a few examples of mounds present on the new block. During the second stage, the counting-by-detection result is used, in addition to block-level information, as the input of the global estimation model to produce a final estimation of the number of mounds. 

\begin{figure*}
    \centering
    \includegraphics[width = 0.9\textwidth]{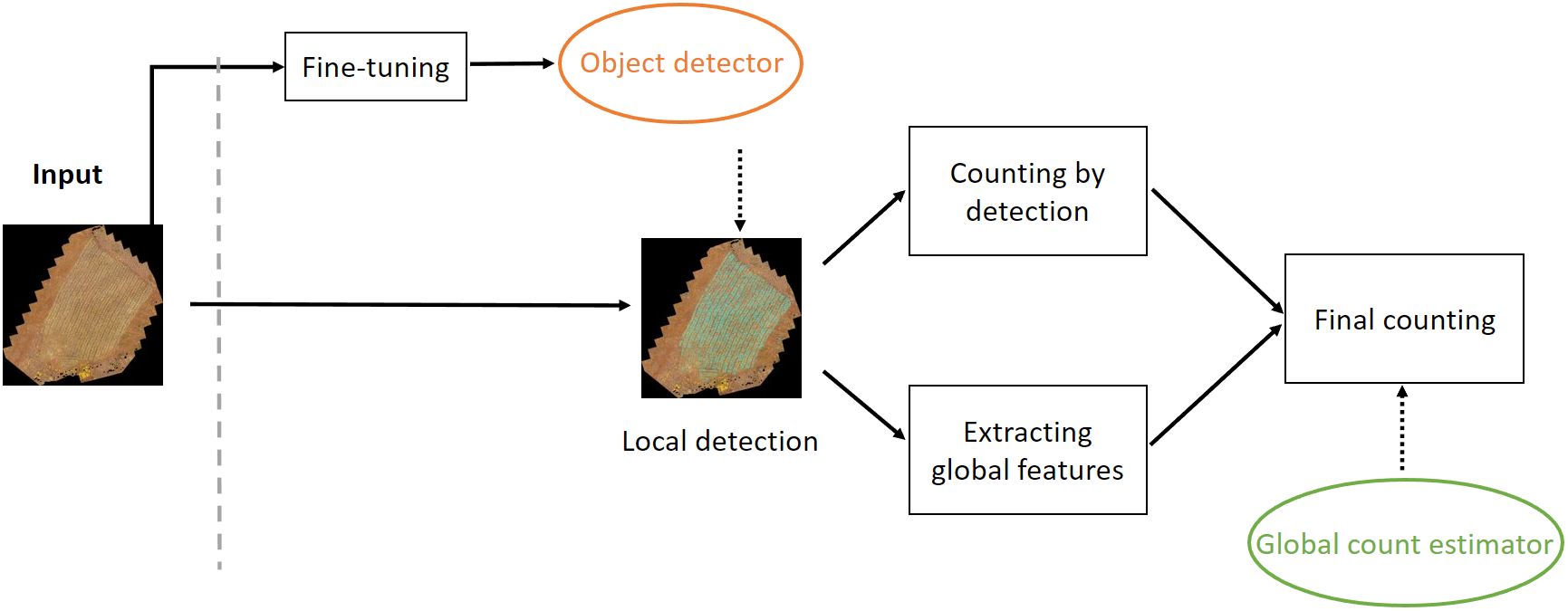}
    \caption{Illustration of our framework while processing new orthomosaic. We fine-tune the trained object detector, then we infer patches to the detector for local mound detection. Once the number of detectable mounds is estimated, a feature vector is constructed by extracting global features. Finally, we apply the global estimator to predict a final count.}
    \label{fig:newimfig}
\end{figure*}

As explained above (and demonstrated further in the experiments), the only use of local object detection is insufficient, as local visual detection may be hampered by various perturbation factors, such as the presence of invisible mounds (e.g. occluded, destroyed) and confusion with the background texture. Both models, each using a different type of feature, are thus essential and complementary for accurate counting under our application constraints. Figure \ref{fig:newimfig} summarizes the process of analyzing a new orthomosaic for mound counting. The proposed framework is detailed in the following sections.
%Motivated by the results achieved, we formulate the count of microsites planting as a detection problem. However, hand crafted features for texture classification are subject of errors when applied to environmental images. Also, region proposals are time and resources consuming when maximizing the search area. Deep Learning object detectors had known a fast evolution and could be classified into two categories: two-stage detectors [18]–[20], and one-stage detectors [21]–[23]. Most advanced one-stage detectors outperform two-stage ones in multiple axes. In one hand, they are faster and more accurate. In the other hand, they are trained in one time and supports transfer learning for specific small dataset applications.

%Figure \ref{fig:trainfig} shows the two training stages of our method. First, Stage1 illustrates the object detector training. Secondly, stage two demonstrate the training stage for the global count estimator.
%The proposed algorithm classifies regions, on each image patch, that have a maximum likelihood of being mound by using a CNN-based detector. Then, the number of detected mounds will be adjusted by regression analysis to find a mapping function between the detected number and the real one.

%In the rest of the section, we present the workflow of our method in three principal steps:  1) base detector construction, 2) Fine tuning, 3) Final count estimation.

\subsection{Local visual detection}

In this work, we use YOLO \cite{Redmon2018YOLOv3AI} as a supervised one-stage approach for object detection. YOLO achieved state-of-the-art performance in several detection tasks while demonstrating robustness against scale and perspective variations \cite{Hamed2021YOLOBB, Hsu2021RatioandScaleAwareYF}. This method is built on the principle of merging very deep CNNs for feature extraction with multi-scale detection in a one-stage detection approach. As a result, the inference time for an image with a size of $608 \times 608$ pixels is about $50 ms$ on a GPU device, with a mean average precision (mAP) of 33\% on COCO dataset \cite{Lin2014MicrosoftCC}. More generally, one-stage object detectors were mainly introduced to decrease the detection time. These detectors are conceptually different from region proposal networks \cite{Ren2015FasterRT} in that they consider the image as a grid of cells that should be scanned only once.

\subsubsection{CNN architecture}

YOLO has a fully convolutional architecture based on Darknet-53 backbone and an output block within 53 convolutional layers for detection on three different scales. The feature extractor mechanism is based on Features Pyramid Network (FPN) architecture \cite{Lin2017FPN} within residual convolutional blocks, and provides feature volume maps as output. YOLO head detector has   $1 \times 1$ convolutional layers to reduce the depth of the volume activation maps without affecting the spatial resolution. Detection is made in three different scales, which are precisely given by down-sampling the dimensions of the input image by 32, 16 and 8 respectively. The advantage of detecting at three scales is to handle in a better way small object detection. Scales are designed to perform small (map size of $52 \times 52$), medium ( $26 \times 26$), and large object detection ($13 \times 13$). An FPN architecture was implemented within the network to perform the map up-sampling process during detection.

The detector analyzes the input image cells to identify objects and produce outputs: position, size, class confidence score, and class number. The position is given by the center location $(C_x, C_y)$, and the size is determined by the width and height of the bounding box $(C_w, C_h)$. The class confidence score is given for each detected object $(C_s)$. For each bounding box, the model classifies the object as belonging to any of the defined classes $C_c$. The output vector is in the form $(C_x, C_y, C_w, C_h, C_s, C_c)$ for each bounding box. YOLO has a very deep fully convolutional network based on FPN architecture, which is appropriate for handling scale variations, perspective variations, partial occlusion, and crowded scenes. Except for scale that is not expected to undergo significant change (due to a fixed flight elevation), all the other difficulties are generally frequent in our application context.
%In our case, we detect only mounds which is a single class detection. The final output vector will be a vector of size $1 \times 6$ and while training the number of convolutional filters at the detection stage will be set to 18. The input size of the model should be fixed both for training and testing, so we adapt the patch size of our dataset to fit the model requirements.
\subsubsection{Training the detector}\label{training}

Training a deep learning object detector from scratch requires a large annotated dataset and significant computational resources. Therefore, we adopt a transfer learning methodology, which consists of retraining a pre-trained model on our dataset. We start the training of our detector, backbone and head detector using weights obtained on a large-scale dataset, such as ImageNet \cite{Russakovsky2015ImageNetLS}, in order to construct a generic mound base-detector. 

Our dataset consists of 18 reconstructed orthomosaics, each representing a different planting block. Since the mound annotation task is tedious and time-consuming, we only annotate six orthomosaics, comprising in total of 9661 objects, to be used as a training set for the mound detector. The training set thus represents 33\% of the entire orthomosaic dataset, while the remaining 67\% are kept for testing. Due to the high resolution of orthomosaics ($23610 \times 18151$), we adopt a patch-based detection approach, where each orthomosaic is divided into non-overlapped patches using a regular grid with fixed cell size. 

%Then, to better exploit the training set performance and avoid over- or underfitting, we perform k-fold cross-validation to evaluate the detector performance. To split the data onto train/test set for each fold, we set a 5 orthomosaic for train and 1 for test and this strategy is better to test the model on unseen data, see algorithm \ref{alg:kfold}. 

%The train dataset is composed of k images, so for each combination we select $(k – 1)$ images for training and the remaining one for validation.
%\subsubsection{Model selection}

%After each training iteration, we infer the test image to the detector and we collect the sum of detected mounds. Object detection metrics, such as average prcision and F1 score, are not %appropriate to our counting problem because they evaluate the performance of the precise detection rather than accurate count. We adopt the relative count error (RE) metric for each image.
%
%The best model is then choosen as the model that have the minimum relative error on the test image, defined by the equation:
%
%\begin{equation}
%    S^* = \underset{argmin}{T\\{t_i}}\frac{Detection_{T\\{ti}} - Ground truth}{Ground truth}
%    \label{eq:BM}
%\end{equation}
%$T$ is the entire set including the 6 partitions, $t_i$ is the $i^th$ partition where $i=1....6$, and $S^*$ is the optimal subset including 5 images.

%We choose the model that can detect the maximum number of mounds in unseen data during the test phase. In other words the best model is the one that has the minimum absolute relative error.

\subsubsection{Data augmentation}\label{sec:augmentation}

\begin{figure}[ht]
    \centering
    \includegraphics[width = 0.25\textwidth]{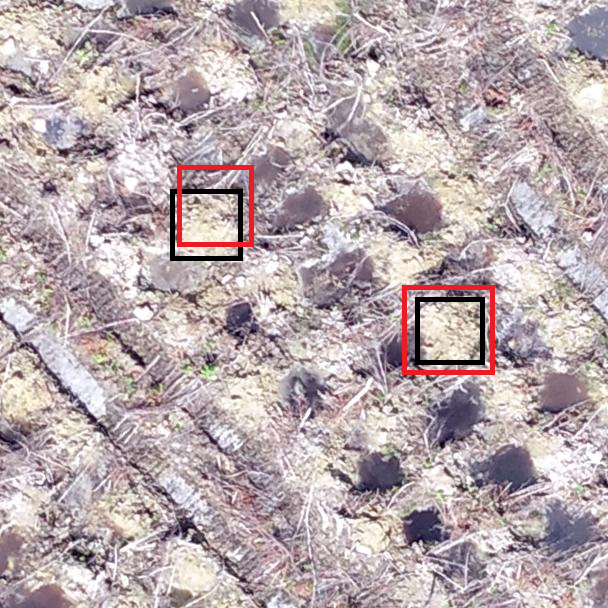}
    \caption{Data augmentation example. Black boxes are the original annotations. Red boxes are examples of generated boxes using the augmentation process.}
    \label{fig:exAug}
\end{figure}

%To construct the base model, we perform two experiments with the objective of verifying the effect of data volume on the model performance. Firstly, we train the model using the annotated data. Secondly, we train the model using an augmented version of the annotated dataset. Our experiments will be included in the cross validation test to validate the best model. 

Data augmentation has been widely used to overcome the lack of data while training a machine learning model. We apply an augmentation process as a series of modifications on initially annotated object patches, to generate additional examples in the training set. In our application, we aim to train the model to detect the presence of mounds without giving much attention to precise localization. This means that we tolerate detection bounding boxes that are not centered on the object. Therefore, in order to generate new training patches, we apply two modifications to bounding boxes to enhance detection performance with respect to background distraction. The applied modifications are 1) size change and 2) translation. Note that these operations are applied to bounding boxes individually, rather than to the entire orthomosaic. We consider that other transformations, like rotation and blurring, are not necessary, since they are sufficiently present in the dataset. On the other hand,  we observed that small-scale changes (caused by differences in sizes between mounds) are efficiently handled by the YOLO architecture described above. Formally, we define our augmentation process by the following operations:
\begin{itemize}
    \item \textbf{Bounding box size change:} we change the scale to include background information from the surrounding region in the bounding box following the equation: 
    
   \begin{equation}\label{eq:trans}
       \begin{cases} H_{new} = H_{old} \times Z \\ W_{new} = W_{old} \times Z \end{cases}
   \end{equation}
         
where $H$ and $W$ are respectively the height and the width of the bounding box, and $Z$ is the scale.
    \item \textbf{Bounding box translation:} we generate non-centered bounding boxes by shifting according to equations: 
    \begin{equation}\label{eq:scale}
        \begin{cases} Cx_{new} = Cx_{old} + ( L \times cos(\alpha)) \\ Cy_{new} = Cy_{old} + ( L \times sin(\alpha)) \end{cases}
    \end{equation}
    
where $Cx$ and $Cy$ are the coordinates of the center of the bounding box, $L$ is the translation amount in pixels and $\alpha$ is the angle defining the translation direction.
\end{itemize}

The overall augmentation process to generate new patches with augmented bounding boxes is presented in the algorithm (\ref{alg:augmentation})
\begin{algorithm}
\SetAlgoLined
\caption{Bounding box augmentation}
\label{alg:augmentation}
\KwData{patch $P$, bounding boxes ${BB}$}
\KwResult{augmented bounding boxes}
\For{$bb \in BB$}{
transform $=$ random(translation, size)\;
\uIf{transform $=$ translation}{
$Z$ = random([0,1])\;
Apply equation (\ref{eq:trans}) using $Z$\;}

\uElseIf{transform $=$ size}{
$L =$ random([0,1])\;
$\alpha$ = random([0,$2\pi$])\;
Apply equation (\ref{eq:scale}) using $(L, \alpha)$\;}
}

\end{algorithm}

\begin{comment}
    The system design of the base detector led to the construction of a generic mound detector from a small dataset. We used transfer learning to train the model and data augmentation to expand the dataset while cross-validation is used to optimize the model accuracy by dividing the data into different sets.
\end{comment}

\subsection{Global count estimation}

In visual object counting, publicly available datasets are generally fully annotated (e.g. Shanghaitech \cite{Zhang2016SingleImageCC}, UCSD \cite{Chan2008PrivacyPC}, UCF\_{Cc} \cite{Idrees2013MultisourceMC}, and UCF-QNRF \cite{Idrees2018CompositionLF}). Therefore, object count generally corresponds to the number of annotated visible objects. However, in our case, the final count is available only after completing planting operations for the corresponding block. Thus, we consider the actual number of planted seedlings as the final count for a given block, which is different from counting manual annotations. 

In other words, our objective is to predict the number of tree seedlings to be planted, which is different from the number of visible mounds. This objective cannot be achieved by the only use of the local visual detector due to two main reasons. First, visual detection of mounds is subject to recognition errors under challenging situations, which may result in both missed detection and false positives. Second, the difference between the detection result and the number of plant seedlings finally used may also be explained by the fact that tree seedlings can be planted in positions where there are no visible mounds (e.g. microsite eroded and collapsed by rainwater, completely occluded by woody debris and coarse rock fragments).

In order to predict the number of required tree seedlings for a given block, we define a global count estimator, which consists of an orthomosaic-level model taking as input global information on the planting block. This model is trained from global information on orthomosaics corresponding to planted blocks  (i.e. for which the number of plant seedlings finally used is known). Given a set of $N$ training observations $X = \{x^i\}, i= 1,2...N$ (representing $N$ planting blocks),  and corresponding ground-truth count $Y= \{y^i\}, i=1,2...N$, for each observation $x^i$ we define a set of $M$ global features $x^i= \{x_j^i\}, j=1,...,M$.

We want to learn a mapping function $F : X \xrightarrow{} Y$, to estimate the final count from the set of global features for each block. We use regression analysis as a predictive technique to model the relationship between predictor variables ($x^i$) and the target variable ($y^i$). We use ridge regression to minimize the loss function (\ref{eq:reg}) and find the M-dimensional weight vector $W= \{w_j\}, j=1,...,M$. The loss function is defined as:

\begin{equation}
    \underset{w}{min}  \sum_{i=1}^{N} \bigg( y_i - \sum_{j=1}^{M} w_j \times x_j^i \bigg)^2\ + \lambda \sum_{j=1}^{M} w_j^2
    \label{eq:reg}
\end{equation}
where $y_i$ is the target variable, $x_j^i$ is the predictor variable, $w_j$ is the weight to be learned, and $\lambda$ the shrinkage parameter controlling the balance between prediction error and regularization of $W$.

To train the regression model, we construct a training dataset where each training example is a global feature vector describing an orthomosaic. Note that prior to constructing the feature vector of an orthomosaic, the local detector is applied in order to obtain a detection-based count as described in section \ref{sec:FT}. In addition to detection-based count, the feature vector $x^i$ includes other block-level information, namely, the average detection density per patch, the average mound density according to user input (i.e. fine-tuning annotations), and the planting block area. 

\subsection{Processing a new orthomosaic}\label{sec:FT}

\begin{figure*}[ht]
   \centering
\subfloat[][]{\includegraphics[width=0.25\textwidth]{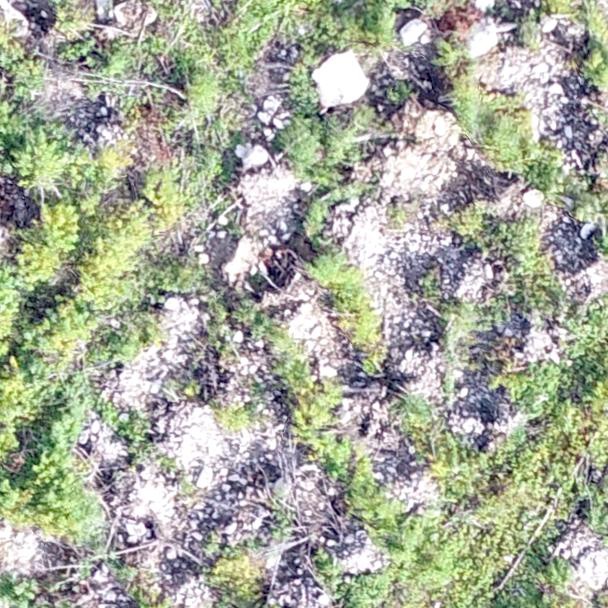}}%
\hfil
\subfloat[][]{\includegraphics[width=0.25\textwidth]{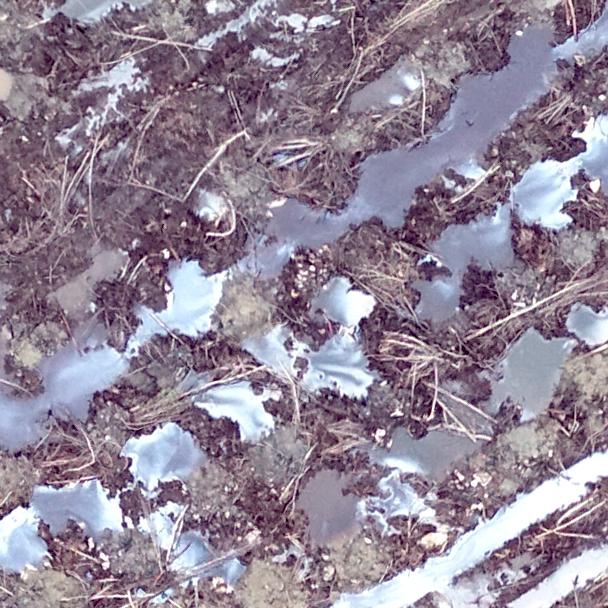}}%
\hfil
\subfloat[][]{\includegraphics[width=0.25\textwidth]{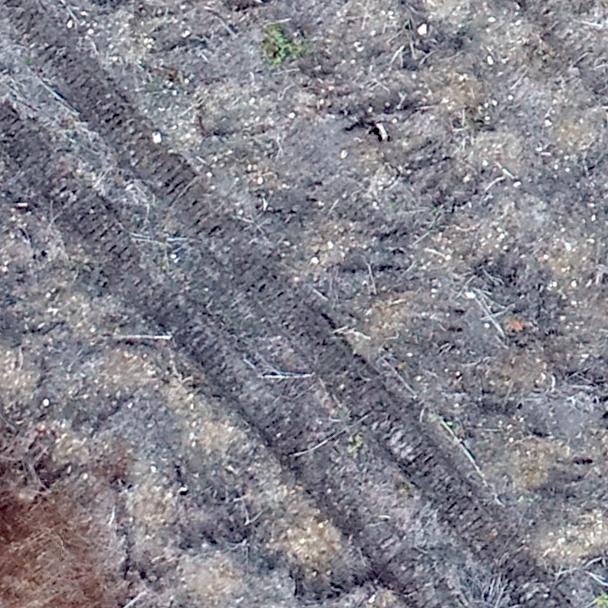}}%
\hfil
    \caption{Three patches from different orthomosaics illustrating inter-block variability, which makes mound identification difficult even for humans. (a) The presence of grass causes partial occlusion of mounds. (B) Water accumulation and mound erosion/deterioration are caused by heavy rain events. (c) Dry terrain where mounds have a similar texture to surrounding regions.}
    \label{fig:FT}
\end{figure*}

The variability of terrains in the dataset results in a high appearance variation for both mounds and surrounding areas. Figure \ref{fig:FT} shows examples of patches from different orthomosaics, illustrating inter-block variability. To address this issue, we fine-tune the local mound detector before processing a new block, in order to adapt the model to the specificities of the new site (e.g., geographic zone, meteorological conditions, season).

The fine-tuning process aims to increase the detector's ability to discriminate mounds from the surrounding area, in a specific context. In our conception, this is achieved by taking model weights obtained by transfer learning (see section \ref{training}), and by retraining only the head of the detector, while the backbone layers are fixed. In summary, context adaptation is performed for a given orthomosaic representing a new planting block through the following three steps:

\begin{enumerate} 
    \item Annotating a small batch of patches from the new planting block,
    \item Generating a batch of images using data augmentation (see Algorithm \ref{alg:augmentation}),
    \item Fine-tuning the visual detector using both annotated and generated images.
\end{enumerate}
The fine-tuned detector is then applied to the new orthomosaic to produce a count-by-detection, which is incorporated with global variables to form the block-level feature vector $F_{i}$ for $x^i$.

To produce the final count, we finally perform a linear transformation of the feature vector using the estimated weight vector $W$:
 \begin{equation}
 \label{eq:finacount}
       %F(i) = W^T x^i = w_0x_0^i + w_1x_1^i + \dots + w_M x_M^i 
       Count_{final}(F_{i}) = \sum_{j=1}^{M} w_j \times x_j^i.
    \end{equation}
Algorithm \ref{alg:finacount} illustrates the main operations performed when processing a new orthomosaic. 

\begin{algorithm}
\SetAlgoLined
\caption{processing new orthomosaic}
\label{alg:finacount}
\KwData{new orthomosaic $T$, trained detector $D$, trained global estimator (equation \ref{eq:finacount})}
\KwResult{Final count}
\Begin{
Annotate a small batch of patches \;
Apply data augmentation (algorithm \ref{alg:augmentation})\;
$D_f$ = Fine-tune $(D)$ \; 
$C_{det}$ = Detect $(D_f, T)$ \; 
Extract global features $F_{ext}$ from $T$ \;
Construct feature vector: $F=(C_{det}, F_{ext})$\; 
Final counting: apply equation \ref{eq:finacount} on $F$  \;illustrate
}
\end{algorithm}

\section{Experiments and results}\label{experim}
\subsection{Dataset construction}

\begin{figure}[ht]
    \centering
    \includegraphics[width = 0.5\textwidth]{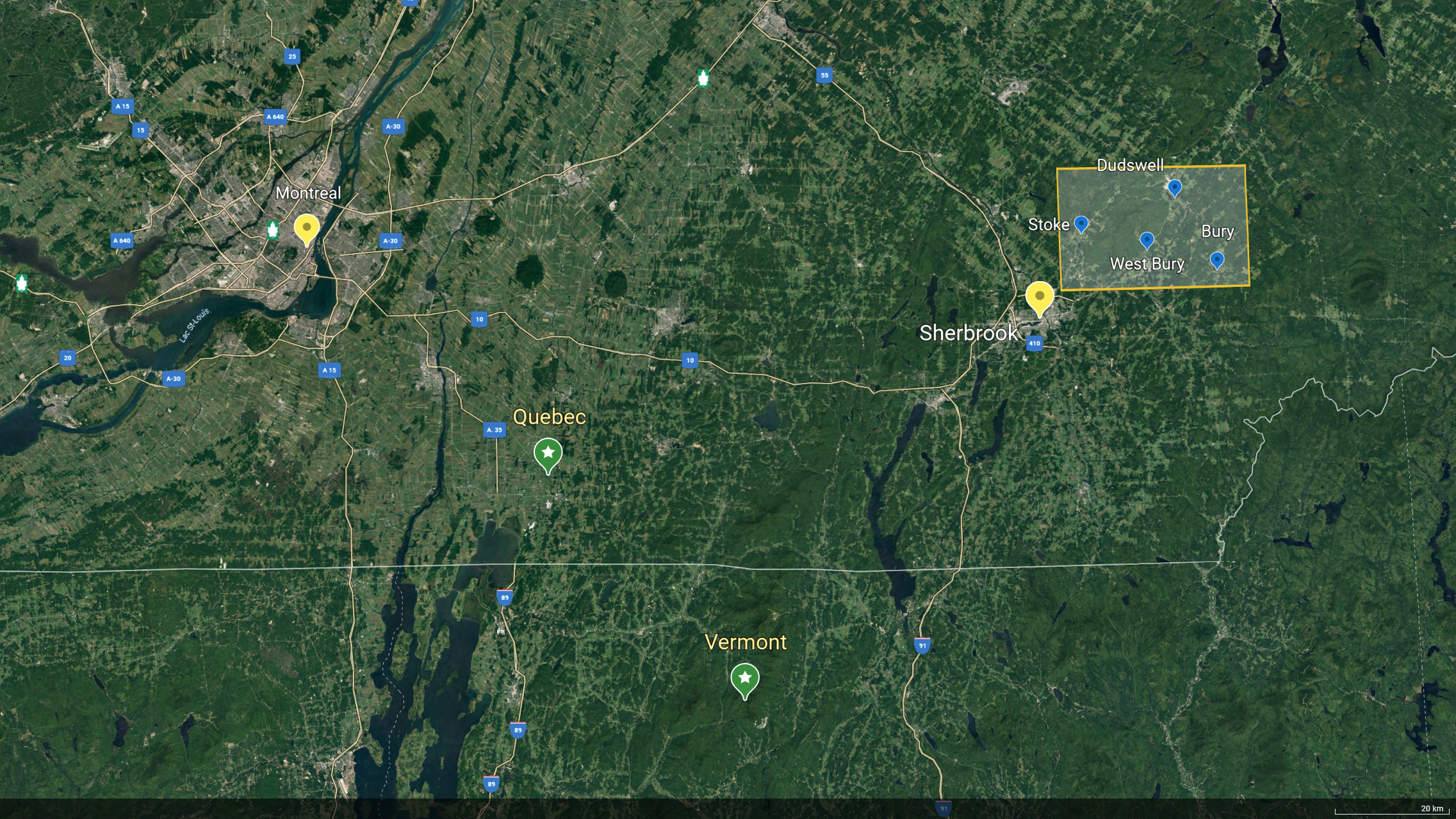}
    \caption{Geographic zone of the study. The yellow rectangle shows the study area in the south of Quebec, Sherbrook city. Map captured from google earth at an elevation of 244 m.}
    \label{fig:geo}
\end{figure}

Figure \ref{fig:geo} shows different sites in the south of the province of Quebec, Canada, where the study has been conducted. Images were captured at an altitude of 120 m. The sensor was set vertically, and images were captured with a high overlap percentage to maximize orthomosaic reconstruction quality. Orthomosaic reconstruction was performed using the Pix4D software. Terrains included in this work are from different zones and have different characteristics. Figure \ref{fig:ortho} shows three different orthomosaics after image reconstruction. A total number of 18 orthomosaics have been reconstructed for 18 planting blocks using the captured images. The number of mounds may vary significantly for different orthomosaics, depending on several factors, such as terrain characteristics and the type of machinery used for mechanical preparation.

\begin{figure}[ht]
   \centering
\subfloat[][]{\includegraphics[width=0.16\textwidth]{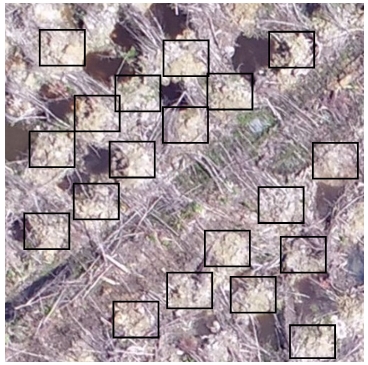}}%
\hfil
\subfloat[][]{\includegraphics[width=0.16\textwidth]{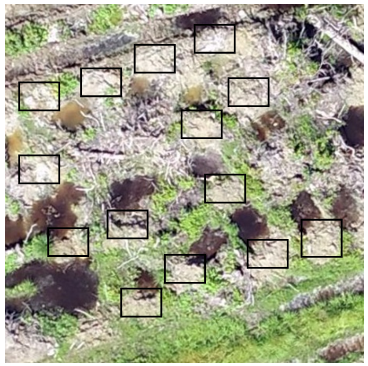}}%
\hfil
\subfloat[][]{\includegraphics[width=0.16\textwidth]{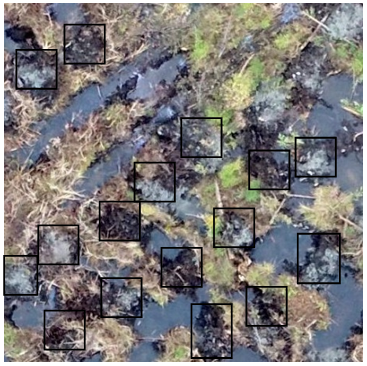}}%
\hfil
    \caption{Three annotated patches belonging to different orthomosaics in G1.}
    \label{fig:annotex}
\end{figure}

We divide our orthomosaic dataset into two groups, G1 and G2.
\begin{itemize}
    \item \textbf{G1:} includes six (6) orthomosaics that we manually annotated for training the object detector and analyzing detection performance. In total, 9661 mounds were manually annotated to train the visual detector. Figure \ref{fig:annotex} shows manual annotations on examples of extracted patches.
    \item \textbf{G2:} comprises twelve (12) orthomosaics that we mainly use for evaluating prediction performance for the entire framework. Note that mounds are not annotated in this dataset, and only global features are used when processing G2.
\end{itemize}

To fit the detector input format, each image is divided into a nonoverlapping regular grid, where the cell (patch) size on the grid is $416 \times 416$ pixels. This division is performed in a manner to maximize the resolution of image patches (cells), which are processed by the object detector. 
%\begin{figure}
%    \centering
%    \includegraphics[width = 0.25\textwidth]{Images/Image3-grid.png}
%    \caption{Example of grid division for one image}
%    \label{fig:grid}
%\end{figure}

\subsection{Evaluation metrics}
To evaluate the visual detection performance, we use precision (P), recall (R), average precision (AP), and F1 score defined as follows:
 \begin{equation}
    P = \frac{TP}{TP + FP}    
    \end{equation}
     \begin{equation}
        R = \frac{TP}{TP + FN}
    \end{equation}
     \begin{equation}
        AP = \sum_{i=0}^{n-1}(R_{i+1} - R_i)P_{interp}(R_{i+1})
    \end{equation}
     \begin{equation}
        F1 =\frac{2TP}{2TP + FP +FN}
    \end{equation}
where $TP$ is the number of True Positives, $FP$ is the number of False Positives, and $FN$ is the number of False Negatives. $P_{interp}(R)$ is the precision interpolated at a certain recall level.
To evaluate the entire framework performance in the counting task, we adopt the relative counting precision metric defined by:
\begin{equation}
    RP = 1-(\|\frac{\# \text{Predicted\_mounds} - \# \text{GT}}{\# \text{GT}}\|)
    \label{eq:RE}
\end{equation}
where $\# \text{Predicted\_mounds}$ is the predicted number of mounds and $\# \text{GT}$ is the ground truth number.

\subsection{Implementation}
The proposed method was implemented using Python on a PC (CPU i7-8700 @ 3.2GHZ, 6 cores) equipped with a GPU Nvidia Geforce GTX 1070. To train YOLOv3, we set the batch size to 16 and the learning rate to 0.001, with a decay of 0.0005. The momentum is set to 0.9 and the number of epochs is fixed to 30 for detector training, and 10 for fine-tuning.

The augmentation parameters $Z$, $L$, and $\alpha$  are respectively set to random numbers in $[0.8, 1.2], [1,10]$, and $[0,2\pi]$. The regressor parameter $\lambda$ is set to 10. For constructing the object detector, we set patch size to  $256 \times 256$ pixels, which produces approximately 10000 training patches after data augmentation, comprising around 95000 annotated mounds for 30 epochs. Transfer learning for YOLO is done based on the weights of the pre-trained model on the ImageNet dataset \cite{Russakovsky2015ImageNetLS}. For fine-tuning, we use a batch of 11 patches of size $256 \times 256$ for each orthomosaic. During the detection process, we set a confidence threshold of 0.25 to detect the most possible mounds without increasing the number of false positives.

Since YOLO implementation was optimized and coded in C, we use the official implementation published by the authors for visual mound detection. We propose a sequential framework to count mounds from orthomosaics.  The input is an orthomosaic representing a planting block.  First, the input orthomosaic is divided into equal patches of size $608 \times 608$ pixels. Then, we annotate a small batch of patches manually to fine-tune the trained visual detector. We fine-tune the model by training the head detector for 10 epochs. After that, the trained model is used to detect mounds in each patch, and results are stored in text files to be used to extract global features. Once visual detection is performed, we construct a global feature vector using annotations and detection results. The total number of the detected mounds is calculated by summing the locally detected mounds on each patch. Then, the average mounds density per patch is calculated for both annotations and detections. Finally, the global feature vector is constructed by adding the planting block area, which is fed to the trained global estimator to predict final count.

The total parameters number of our model is $62,001,757$. The visual object detector inference speed is equal to 45 patches per second. The total time required by our framework to predict the final count depends on the orthomosaic size, which vary between $10000 \times 15000$ and $20000 \times 60000$ pixels, depending on camera settings and planting block area.

\subsection{Ablation study}
\subsubsection{Local visual detector}
\label{sec:CVdetector}

In this section, we present an ablation analysis on the $G1$ subset to examine the impact of data augmentation and fine-tuning on detection performance. We evaluate three different scenarios by training and testing three versions of the object detector :
\begin{itemize}
    \item \textbf{Model 1:} We do not use data augmentation so that only annotated data are used for training the detector. For detection, the trained model is applied directly without fine-tuning.  
    \item \textbf{Model 2:} We use data augmentation but not fine-tuning.
    \item \textbf{Model 3:} We use the complete version of our detector as described in section \ref{training}, including data augmentation and fine-tuning steps.
\end{itemize}

 Our evaluation is based on a cross-validation methodology, which performs a circular combination among all possibilities for training and testing. Since the amount of annotated data (6 orthomosaics) is limited compared to the entire dataset (18 orthomosaics), cross-validation is beneficial in that it allows to efficiently exploit available annotations, while mitigating the data split effect  \cite{Stone1974CrossValidatoryCA, Fushiki2011EstimationOP, Kopper2020ModelSA}.
 
 In each of the six iterations, we partition data as follows.
\begin{itemize}
    \item Five (5) orthomosaics are used for training/validation. The obtained patches are shuffled and then split into two subsets for training (90\%) and validation (10\%).
    \item One (1) orthomosaic is not seen during training and is used only for testing.
\end{itemize} 
The cross-validation process is illustrated in algorithm \ref{alg:kfold}.

%faire ref a l'algo
\renewcommand{\arraystretch}{2}
\begin{table*}[]
    \centering
    \caption{Detection accuracy results for the local object detector trained on six different folds.}
    \begin{tabular}{|c|c|c|c|c|c|c|c|c|c|c|c|r|}
    
    \hline
  \multirow{2}{*}{Fold} & \multicolumn{4}{c|}{Model1}&  \multicolumn{4}{c|}{Model2} & \multicolumn{4}{c|}{Model3} \\
   \cline{2-13} 
    & P & R & AP & F1 & P & R & AP & F1& P & R & AP & F1\\
    \hline
    1 & 38.4\% & 16.4\% & 10.7\% & 23\% & 39.3\% & 19.3\% & 16.9\% & 25.9\% & 30\% & 21.4\% & 12.9\% & 25\%\\
    2 & 44.7\% & 39.6\% & 28.3\% & 42\% & 41.4\% & 40.7\% & 32.5\% & 41.1\% & 41.4\% & 40.7\% & 32.5\% & 41.1\%\\
    3 & 61.7\% & 10.9\% & 14.7\% & 18.5\% & 58.3\% & 12.1\% & 16\% & 20\% & 23.7\% & 21.8\% & 11.1\% & 0.227 \\
    4 & 34.7\% & 21.1\% & 14.3\% & 26.2\% & 27.4\% & 22.9\% & 14.3\% & 25\% & 37.9\% & 62.5\% & 40.9\% & 47.2\% \\
    5 & 35.1\% & 20.7\% & 16.8\% & 26.1\% & 58.9\% & 25.1\% & 24.1\% & 34.2\% & 32.4\% & 36.9\% & 28.2\% & 0.345 \\
    6 & 49.2\% & 29.8\% & 19.8\% & 37.1\% & 35.5\% & 40.2\% & 23.5\% & 37.7\% & 31.9\% & 47.5\% & 28.9\% & 38.2\% \\
    \hline
    Average & 43.96\% & 23.08\% & 17.43\% & 28.82\% & 43.47\% & 26.72\% & 21.22\% & 30.65\% & 32.88\% & 38.47\% & \textbf{25.75}\% & \textbf{34.78}\%  \\
    \hline
    \end{tabular}

    \label{tab:resG1}
\end{table*}

\begin{algorithm}
\SetAlgoLined
\caption{K-fold cross validation}\label{alg:kfold}
\KwData{$K$ orthomosaics represented by $T=\{T_i\}$, ground-truth counts  $Y=\{Y_i\}$}
\KwResult{Average error over tests}
\For{$i \in \{1,...,K\}$}{
    Training\_set = $\{T\backslash T_i\}$ \; 
    Testing\_set = $T_i$ \; 
    Detector = Train (Training\_set) \; 
    $h_i$ = Detect (Detector, Testing\_set) \; 
    $e_i$ = Calculate\_error ($h_i, Y_i$) \;
    }
 \textbf{return}  $\frac{1}{k} \sum_{i=1}^{k}(e_i)$
\end{algorithm}

Table \ref{tab:resG1} shows results of the 6-fold cross-validation for the three detector versions. We report precision (P), recall (R), average precision (AP), and F1 scores for all experiments. Average scores over all iterations (mainly AP and F1) show that training the detector using augmented data enhances recognition performance compared to the first version, where only annotated data are used. This suggests that using only annotated data may cause under-fitting due to the lack of training examples. Thus, the data augmentation process allowed to improve detection performance, which resulted in increasing AP and F1 scores by 3.8\% and 1.8\%, respectively. However, the best performance over the three scenarios was obtained by using the complete version of the visual detector, which achieved an average detection precision of 25.75\% and an F1 score of 34.78\%. This result demonstrates the importance of data augmentation during training and fine-tuning when applying the model to a new image.  

\begin{figure*}[ht]
   \centering
\subfloat[][]{\includegraphics[width=0.25\textwidth]{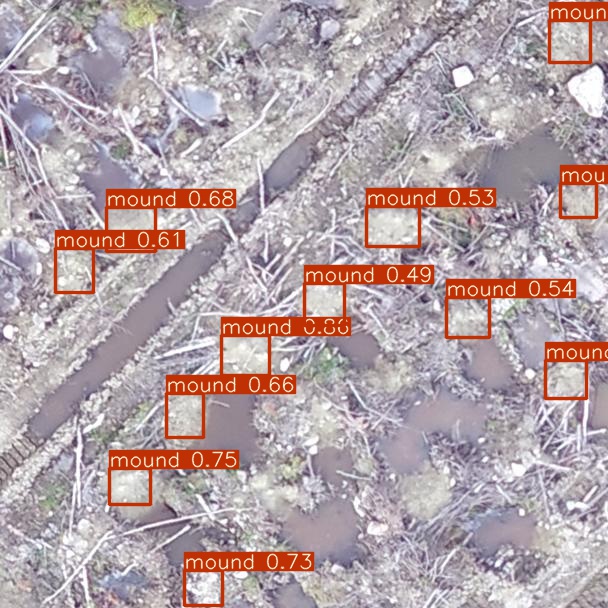}}%
\hfil
\subfloat[][]{\includegraphics[width=0.25\textwidth]{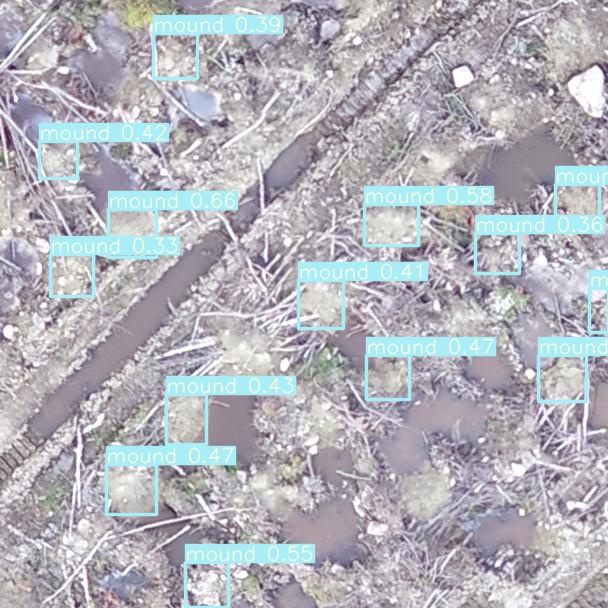}}%
\hfil
\subfloat[][]{\includegraphics[width=0.25\textwidth]{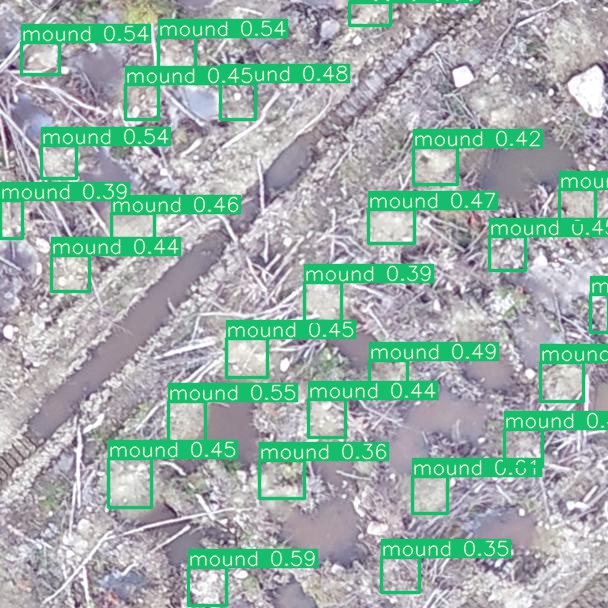}}%
\hfil
    \caption{Qualitative results for the three versions of the object detection model on the same patch comprising 24 annotated mounds. The number of detected mounds is 12 for Model 1 (a), 13 for Model 2 (b), and 24 for Model 3 (c).}
    \label{fig:resDetG1}
\end{figure*}

In Figure \ref{fig:resDetG1}, we report qualitative results of the three detectors when processing the same image patch of size $498 \times 498$, containing 24 objects, according to manual annotations. This example shows that the final version of the detector has a better ability to recognize objects of interest. In fact, augmenting the training set and fine-tuning the detector allowed to increase the detector's ability to discriminate objects from background regions. Moreover, the context information incorporated during fine-tuning is shown to be beneficial in adapting the model to specific situations.

\subsubsection{Count correction using global features}
\label{sec:ablation.global}
To evaluate the contribution of the global correction module in the final prediction, we perform another ablation analysis on the $G1$ dataset. This is done by comparing the counting relative precision of the local visual detector to that of the entire system, including both local detection and global correction process.

\renewcommand{\arraystretch}{1.5}
\begin{table*}[]
    \centering
    \caption{Relative counting precision of two different versions of the system. Testing is done according to the same cross-validation methodology described in section \ref{sec:CVdetector}. For each cross-validation iteration, testing is performed on an orthomosaic representing a plantation block.}
    \begin{tabular}{|c|c|c|c|c|c|c|}
    \hline
      \multirow{2}{*}{Orthomosaic}   & \multirow{2}{*}{GT} & \multicolumn{2}{c|}{Detection-based count} & \multicolumn{3}{c|}{Globally corrected prediction}\\ 
        \cline{3-7}
        &  & Count & RP & Count & RP& Improvement\\
        \hline
    T1  & 3750 & 1911 & 51\% & 3580 & \textbf{95.5}\%& \textbf{45.5}\%\\    
       \hline
    T2  & 1450 & 1074 & 74.1\% & 1166 & \textbf{80.4}\%&\textbf{6.3}\%\\  
       \hline
    T3  & 1350 & 836 & \textbf{61.9}\% & 2004 &  51.6\%& -10.3\%\\  
       \hline
    T4 & 4750 & 5407 & 86.2\% & 5051 & \textbf{93.7}\%&\textbf{7.5}\%\\
    \hline
    T5 & 8600 & 6196 & 72.0\% & 8334 & \textbf{96.9}\%&\textbf{24.9}\%\\
    \hline
    T6 & 6500 & 6866 & 94.4\% & 6172&\textbf{95}\%&\textbf{0.6}\% \\
    \hline
    Average &\multicolumn{2}{c|}{}  & 73.2\% & & \textbf{85.4}\%&\textbf{12.4}\%\\
    \hline
    \end{tabular}
    
    \label{tab:resG1re}
\end{table*}

Table \ref{tab:resG1re} presents the counting results of two different versions of our framework. The first version consists of the object detection module used for predicting detection-based count, while the second consists of the entire framework including both detection and correction procedures as illustrated in Figure \ref{fig:exAug}. Results are reported for the $G1$ subset using LOOCV on six different planting blocks. From table \ref{tab:resG1re}, we can see that the global correction function improved counting precision for five over six planting blocks, with an average improvement of $12.4\%$. For example, This improvement reached $45,5\%$ for planting block T1. From this experiment, we conclude that counting-by-detection is insufficient to provide precise counting under our application constraints and that global prediction is essential to perform counting correction based on global features of planting blocks.

\subsection{Real-world application results}

\renewcommand{\arraystretch}{1.5}

\begin{table*}[]
    \centering
    %%\textcolor{blue}{
    \caption{Quantitative results of our proposed method. GT is the ground-truth corresponding to the number of plant seedlings planted in a block. The count is the predicted number of mounds and RP is the relative counting precision. We report results for detection-based prediction using YOLO \cite{Redmon2018YOLOv3AI} (without global correction), globally corrected prediction (detection followed by correction), and Faster RCNN \cite{Ren2015FasterRT}. The average precision measure represents the average over all precision values, while overall precision stands for the counting precision considering the total number of mounds in the dataset. Note that planting blocks with a larger number of mounds have more significant contributions to the overall precision.}
    \begin{tabular}{|c|c|c|c|c|c|c|c|}
    \hline
      \multirow{2}{*}{Orthomosaic}   & \multirow{2}{*}{GT} & \multicolumn{2}{c|}{YOLO \cite{Redmon2018YOLOv3AI}} & \multicolumn{2}{c|}{Faster RCNN \cite{Ren2015FasterRT}} & \multicolumn{2}{c|}{Ours} \\ 
        \cline{3-8}
        &  & Count & RP & Count & RP & Count & RP \\
        \hline
    T7   & 8900 & 7713 & 86.7\%& 4441 & 50\%& 8318 & \textbf{93.5\%}\\    
       \hline
    T8  & 1400 & 1023 & 73.1\% & 1058 & 76\%& 1734 & \textbf{76.1\%}\\  
       \hline
    T9  &  12350 & 8442 &  68.4\%& 5702 & 46\%& 11773 &  \textbf{95.3\%}\\  
       \hline
    T10 & 16450 &  15661 &  \textbf{95.2\%} & 13467 & 82\%& 15226 &  92.6\%\\
    \hline
    T11 & 3750 &  3214 & 85.7\%& 1586 & 42\%& 4158 & \textbf{89.1\%} \\
    \hline
    T12 & 7150 & 2775 &  38.8\% & 3106 & 43\%& 6022 &  \textbf{84.2\%}\\
    \hline
    T13 & 2300 & 1735 &  75.4\% & 803& \textbf{35\%}& 1775 &  77.2\%\\
    \hline
    T14 & 6600 & 1803 &  27.3\% & 2607 & \textbf{40\%}& 6050 &  91.7\%\\
    \hline
    T15 & 4350 & 2506 &  57.6\% & 132 & 3\%& 3781 & \textbf{86.9\%}\\
    \hline
    T16 & 6150 & 5829 &  94\%& 2354 & 38\%& 6019 &  \textbf{97.9\%}\\
    \hline
    T17 & 2050 & 771 &  37.6 \%& 930 & 45\%& 1764 &  \textbf{86.1\%}\\
    \hline
    T18 & 13100 & 8914 &  68\%& 4572 & 35\%& 14364 &  \textbf{90.3\%}\\
    \hline
     Overall result & 84550 & 60386 & 71.4\% & 58633 &52.8\%& 80989 & \textbf{95.8}\%\\
    \hline
     Average precision & \multicolumn{2}{c|}{} & 67.4\%&  &50.7\% & & \textbf{88.4}\%\\
    \hline
    \end{tabular}
    \label{tab:resG2fe}
     %% }
\end{table*}

In real-world conditions, the process of estimating the number of mounds on a new planting block can be summarized as follows:
\begin{enumerate}
    \item Dividing the input orthomosaic into patches according to a regular grid,
    \item Fine-tuning the object detector using a batch of patches,
    \item Applying the fine-tuned detector and performing counting-by-detection,
    \item Extracting the global feature vector;
    \item Predicting the final count using the trained regressor (equation \ref{eq:reg}).
\end{enumerate}

 Note that our training dataset for the local object detector is composed of the six annotated orthomosaics of G1, which are only used for training. Final testing is performed on the remaining 12 orthomosaics of G2 (from T7 to T18). Since we have a limited training set for the regressor (global estimation function), we adopt a leave-one-out cross-validation-like strategy where the regression training set changes for each testing iteration. That is, for the $i^{th}$ test on orthomosaic $T_i$, the global estimation function is trained on 17 global feature vectors, representing the entire dataset (G1 and G2), except $T_i$. The main motivation for this choice is to fully take advantage of the limited amount of data for which global feature vectors are available.

It can be observed from table \ref{tab:resG2fe} that the comparison between detection-based-counts and globally corrected predictions is consistent with the ablation analysis in \ref{sec:ablation.global}. This result confirms the importance of combining local visual detection with global count estimation. From detailed results, we can also see that the achieved counting precision is superior to that of the method currently used by forest managers (85\%) for nine testing blocks among 12. Moreover, the precision exceeds 90\% for six testing blocks, and our framework outperforms the manual method in both average precision (88,4\%) and overall precision (95,8\%).

We also present a comparison with the state-of-the-art method Faster RCNN \cite{Ren2015FasterRT}, which is a two-stage object detector that uses a Region Proposal Network (RPN) to generate proposals. It is clearly seen from table \ref{tab:resG2fe} that our proposed framework achieved better results. Overall, our method improves the performance by 24\%,  with a 95.8\% overall precisioncompared to 52.8\% for Faster RCNN. In average precision, which reflects accuracy per block, our method is significantly more effective and achieves 88.4\% compared to 67\%, and 50\% for YOLO and Faster RCNN respectively.  

 To further validate the importance of our correction module, we conduct a statistical validation student test, which is a statistical hypothesis test used for small set sizes. We claim that the correction module improves the global count and reduces counting error per block. Thus, we define the null $H_0$ and alternative $H_1$ hypothesis.
\[
     \begin{cases}
     H_0: \mu_{detection} =  \mu_{correction}  \\
     H_1: \mu_{detection}  \leq  \mu_{correction} 
 \end{cases}  
\]
To calculate the p-value, we use the RP calculated over 12 orthomosaic test data using detection and corrected count. Our calculated p-value is equal to $0.00708$ and the statistic value is equal to $-3.299$. T-test result supports our claim that the correction module improves the final count results. At an alpha value equal to $0.05$, the calculated p-value is much lower than the alpha value, which means that the difference between means is statistically significant and the null hypothesis is rejected.

\section{Discussion}\label{disc}

In this section, we discuss the main challenges related to object detection and final count estimation, based on the obtained results.

\subsection{Local object detection challenges} 
\label{sec:local}
The first step in our framework is to infer the orthomosaic to the detector. From table \ref{tab:resG2fe}, we see that for T12 and T14, the counting-by-detection precision is lower than 40\%, as the detector was unable to efficiently recognize mounds in corresponding image patches. This can be explained by the relatively limited amount of annotated data available for training the detector (six blocks), compared to the testing set (12 blocks), which may limit the generalization ability of the detection model. In our experiments, an unseen data example is defined as a new plantation block completely excluded from the training set. Therefore, when testing the object detector, it is possible that the new plantation block shows different unseen characteristics. This may include terrain properties and mound shapes that were not encountered by the detector during training. 

In order to increase the accuracy of the object detector, the optimal data partitioning strategy would have been to include a sample from each of the 18 blocks in the training set. However, we finally defined a more rigorous evaluation strategy, that is more consistent with our real-world problem and application constraints. To maximize the generalization ability of the model when deploying our application, we recommend training the base detector on all available orthomosaics, or at least including a sample from each available orthomosaic in the training set. We also recommend continuously feeding the system with new data as images and annotations become available, in order to continuously improve detection performance.

Note that even with the use of a larger training dataset, several challenging situations may still significantly impact local detection performance due to recognition perturbation factors. These factors mainly include mound occlusion, destruction, and image acquisition issues. 

\begin{figure}[ht]
   \centering
\subfloat[][]{\includegraphics[width=0.2\textwidth]{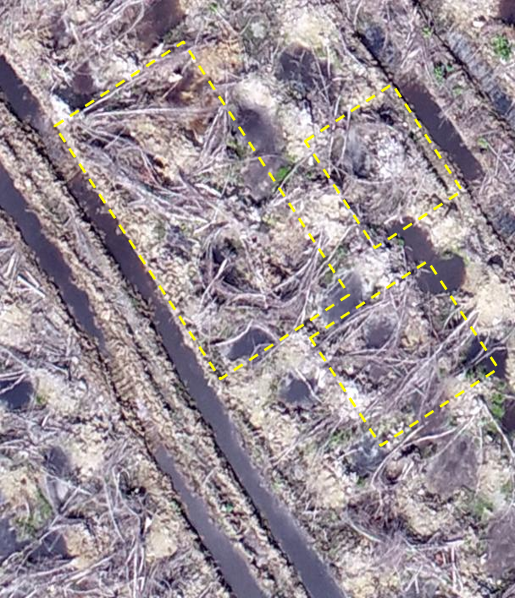}}%
\hfil
\subfloat[][]{\includegraphics[width=0.2\textwidth]{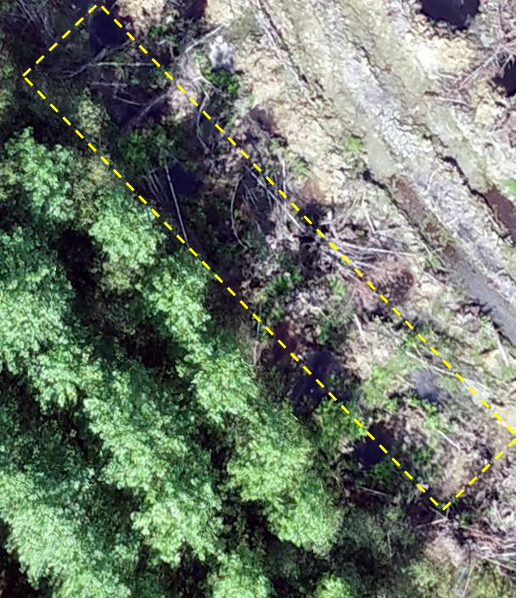}}%
\hfil

    \caption{Examples of occlusion situations on RGB images captured at an altitude of 120 m. (a) Occlusion is due to the presence of woody debris and rock fragments on the forest floor. (b) Occlusion is due to the presence of trees and shadows on border regions.}
    \label{fig:occ}
\end{figure}

\textbf{Occlusion:} As illustrated in figure \ref{fig:occ}.a, this situation may occur due to the presence of woody debris and coarse rock fragments on the forest floor. Moreover, mounds located on block borders may also be occluded by neighboring trees or shadows (see figure \ref{fig:occ}.(b)).

\begin{figure}[ht]
   \centering
\subfloat[][]{\includegraphics[width=0.16\textwidth]{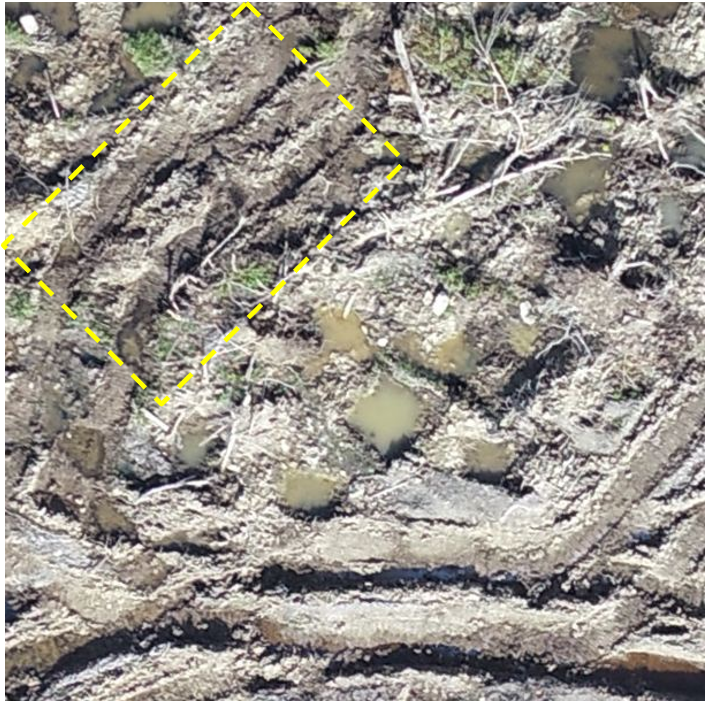}}%
\hfil
\subfloat[][]{\includegraphics[width=0.16\textwidth]{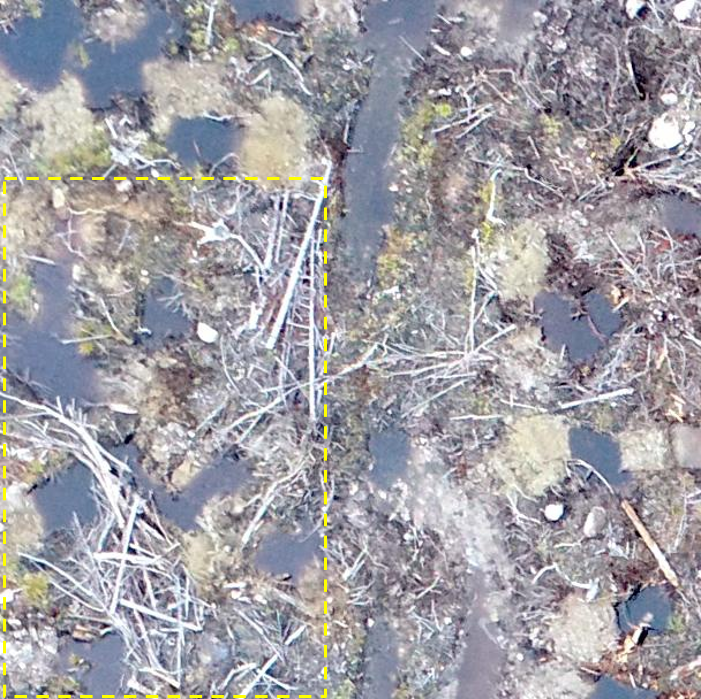}}%
\hfil
\subfloat[][]{\includegraphics[width=0.16\textwidth]{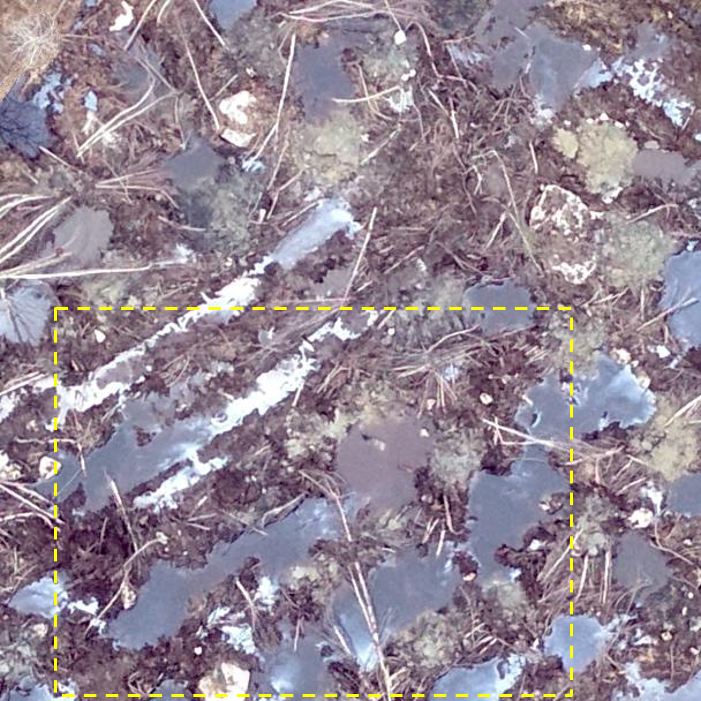}}%
\hfil
    \caption{Examples of aerial RGB images showing destructed mounds. (a) Mounds destroyed mounds by the excavator during mechanical preparation. (b) \& (c) Mound destruction due to heavy rain events. }
    \label{fig:dest}
\end{figure}

\textbf{Mound destruction:} During mechanical preparation of a planting block, mound destruction may be caused by the tracks of the excavator, as shown in figure \ref{fig:dest}(a). Moreover, in the case of heavy rain events following mechanical preparation, the created mounds may be subject to deterioration and erosion. This is because, in intensive silviculture, planting blocks are often along hillslopes, which favors water flow. Figure \ref{fig:dest}(b) \& (c) shows examples of regions affected by these conditions. In order to handle the effects of hydroclimatic conditions, we recommend performing image capture flights as soon as mechanical preparation is complete.

\begin{figure}[ht]
   \centering
\subfloat[][]{\includegraphics[width=0.16\textwidth]{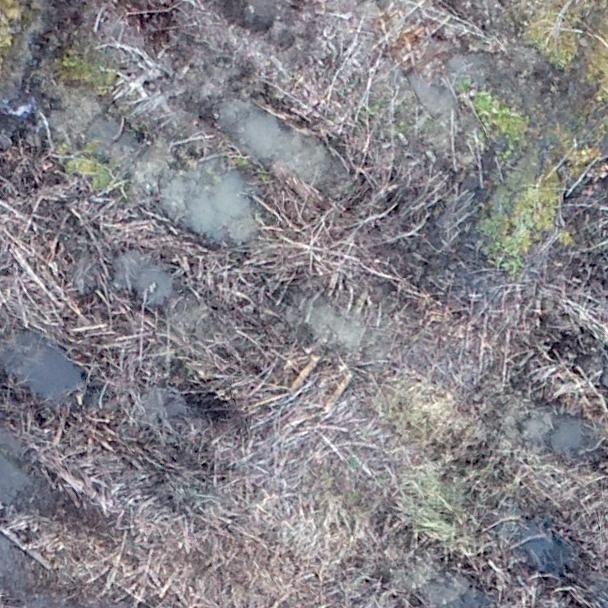}}%
\hfil
\subfloat[][]{\includegraphics[width=0.16\textwidth]{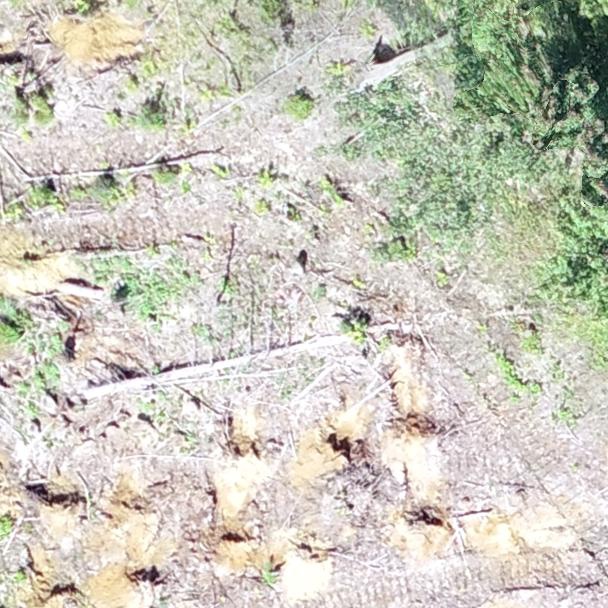}}%
\hfil
\subfloat[][]{\includegraphics[width=0.16\textwidth]{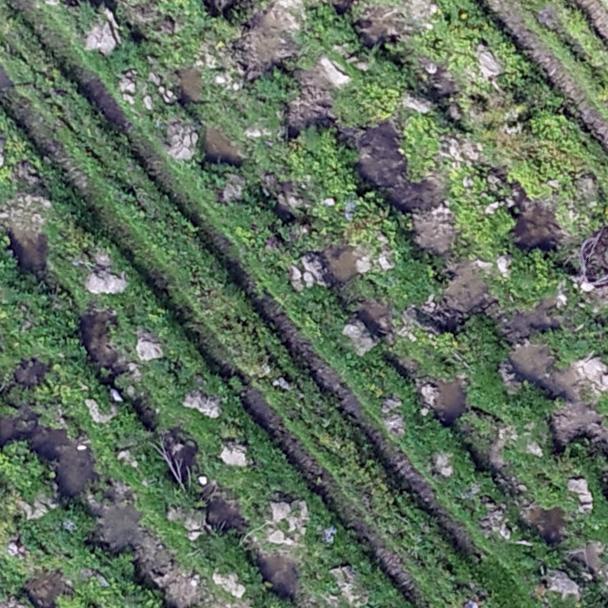}}%
\hfil
    \caption{Examples of visual effects caused by image acquisition conditions. (a) The image is totally blurred. Sunlight reflection increases image brightness in (b), while in (c) the image is relatively dark due to a lack of luminosity.}
    \label{fig:aqui}
\end{figure}

\textbf{Image acquisition factors:} The image acquisition process is an important step that may significantly impact image quality, as well as the appearance of the object of interest. This is mainly due to several perturbation factors that may occur during flights due to weather conditions, such as camera movements due to the presence of wind and lighting variation. These factors may result in several visual effects on the captured images, like blurred image regions and unbalanced colors (see figure \ref{fig:aqui}). 
Object visibility is also related to flight altitude, as flying at a low altitude increases the level of detail in images. However, a good trade-off should be found since on the other hand, flying at a high altitude allows to simplify image acquisition and orthomosaic construction, prior to applying the proposed framework.    

\subsection{Global prediction performance}
The performed experiments show the importance of combining local object detection with a global correction function, in order to improve count estimation performance. This demonstrates that global count estimation based on block-level features (including detection-based count) allows for mitigating the object detection issues discussed above. However, we observed relatively high counting errors during our final tests for plantation blocks  T8 and T13 (see table \ref{tab:resG2fe}). We can notice that these two plantation blocks have small areas (2.37 ha and 3.09 ha, respectively), with a small number of mounds (1400 and 2300, respectively). This performance gap for small area blocks can be explained by two reasons. The first is training data imbalance with respect to the area criteria, as small area blocks are underrepresented in the regressor training set. Second, this can also be explained by block border effects, which are generally more impactful with small blocks (see section \ref{sec:local} and figure \ref{fig:occ}.b). It would be therefore important to feed the global prediction function with additional global observations of small blocks as soon as they become available, in order to overcome the under-representation issue.

\section{Conclusion}\label{concl}

We presented a new computer vision framework to automate the counting process of planting microsites on a mechanically prepared planting block. The proposed system is based on a hybrid approach combining local patch-level detection with global prediction at the block level. The performed experiments demonstrate the effectiveness of our design in handling several challenging situations related to environmental and image acquisition conditions, as well as to inherent limitations of the detection model. Indeed, while the object detection model exploits visual information from UAV images to predict a preliminary count, the subsequent global estimation procedure exploits complementary block-level information to mitigate detection limitations. Furthermore, the proposed solution outperforms traditional counting methods, while offering significant advantages in terms of fieldwork conditions and cost optimization.

Our future work comprises two main lines of research. At the fundamental level, we aim to incorporate both local detection and global correction mechanisms in a single end-to-end deep learning model to address the counting problem. At the application level, we aim at generalizing the model to other computer vision applications implying object counting in crowded scenes.


%\section*{Acknowledgment}

%This work was supported by a Research Grant from the Natural Sciences and Engineering Research Council of Canada (NSERC) [Discovery Grant number RGPIN-2020-%04937]. The authors would like to thank their collaborators from Domtar. The authors would also like to thank the associate editor and the reviewers for their %comments and suggestions that helped to improve the paper.
 
%\section{Simple References}
%You can manually copy in the resultant .bbl file and set second argument of $\backslash${\tt{begin}} to the number of references
% (used to reserve space for the reference number labels box).

%\begin{thebibliography}{1}
\bibliographystyle{IEEEtran}
\bibliography{IEEEabrv, bibtex/bib/Mybib}

%\end{thebibliography}

%\newpage



%\vspace{-33pt}
\begin{IEEEbiography}[{\includegraphics[width=1in,height=1.25in,clip,keepaspectratio]{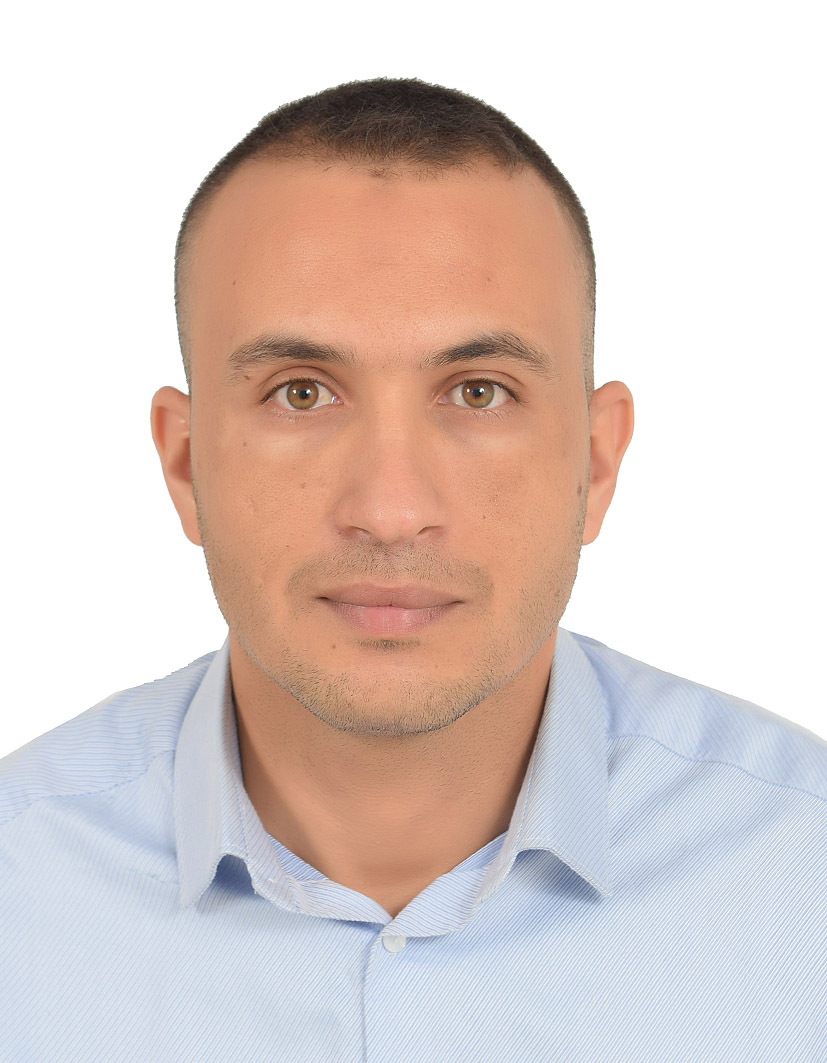}}]{Ahmed Zgaren}
is a Ph.D. student in the Department of Information Systems and Engineering at Concordia University (Montréal, Canada). He holds the Master degree in computer science from the National Engineering School of Tunis (Tunisia),  and the engineer degree from the Military College of Tunisia.  His research interests include computer vision, object detection/tracking, UAV imaging, and speech analysis.
\end{IEEEbiography}

\vspace{11pt}
\vspace{-33pt}
\begin{IEEEbiography}[{\includegraphics[width=1in,height=1.25in,clip,keepaspectratio]{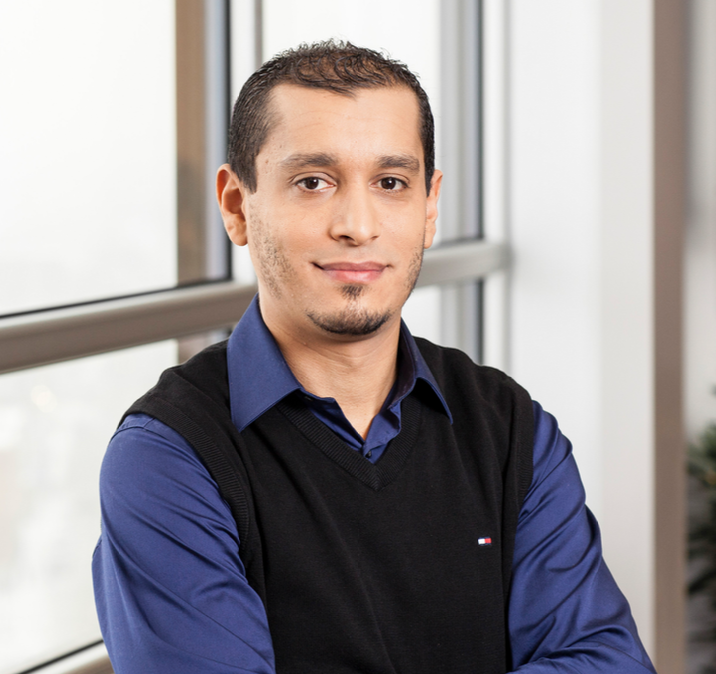}}]{Wassim Bouachir}
is currently a professor of computer science at the University of Québec (TÉLUQ). He holds a Ph.D. degree in computer engineering from Polytechnique Montréal and a M.Sc. in computer science from the University of Moncton. His research interests include fundamental problems in computer vision, signal processing, and machine learning. His research activities also aim to develop AI-based systems for several application areas, such as security, physical and mental health, and environment applications.
\end{IEEEbiography}

\vspace{11pt}

\vspace{-33pt}
\begin{IEEEbiography}[{\includegraphics[width=1in,height=1.25in,clip,keepaspectratio]{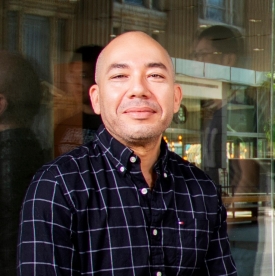}}]{Nizar Bouguila}
 (Senior member IEEE) received the engineer degree from the University of Tunis, Tunis, Tunisia, in 2000, and the M.Sc. and Ph.D. degrees in computer science from Sherbrooke University, Sherbrooke, QC, Canada, in 2002 and 2006, respectively. He is currently a Professor with the Concordia Institute for Information Systems Engineering (CIISE) at Concordia University, Montreal, Quebec, Canada. His research interests include image processing, machine learning, data mining,, computer vision, and pattern recognition. Prof. Bouguila received the best Ph.D Thesis Award in Engineering and Natural Sciences from Sherbrooke University in 2007. He was awarded the prestigious Prix d’excellence de l’association des doyens des etudes superieures au Quebec (best Ph.D Thesis Award in Engineering and Natural Sciences in Quebec), and was a runner-up for the prestigious NSERC doctoral prize. He was the holder of a Concordia University research Chair Tier 2 from 2014 to 2019 and was named Concordia University research Fellow in 2020. He is the author or co-author of more than 400 publications in several prestigious journals and conferences. He is a regular reviewer for many international journals and serving as associate editor for several journals such as Pattern Recognition journal. Dr. Bouguila is a licensed Professional Engineer registered in Ontario, and a Senior Member of the IEEE.
\end{IEEEbiography}

\vspace{11pt}


\vfill

\end{document}